\documentclass[pdflatex,sn-mathphys-num]{sn-jnl}
\usepackage{graphicx}%
\usepackage{multirow}%
\usepackage{amsmath,amssymb,amsfonts}%
\usepackage{amsthm}%
\usepackage{mathrsfs}%
\usepackage[title]{appendix}%
\usepackage{xcolor}%
\usepackage{textcomp}%
\usepackage{manyfoot}%
\usepackage{booktabs}%
\usepackage{subcaption}
\usepackage{algorithm}%
\usepackage{algorithmicx}%
\usepackage{algpseudocode}%
\usepackage{listings}%
\usepackage{float}
\theoremstyle{thmstyleone}%
\theoremstyle{thmstyletwo}%
\theoremstyle{thmstylethree}%
\begin{document}

\title[Article Title]{Training a Custom CNN on Five Heterogeneous Image Datasets}

\author[1]{\fnm{Anika} \sur{Tabassum}}\email{anika-2019417844@cs.du.ac.bd}
\author[1]{\fnm{Tasnuva Mahazbin} \sur{Tuba}}\email{tasnuvamahazabin-2018725348@cs.du.ac.bd}
\author[1]{\fnm{Nafisa} \sur{Naznin}}\email{nafisanaznin24cse041@gmail.com}

\affil[1]{\orgdiv{Department of Computer Science and Engineering}, \orgname{University of Dhaka}, \orgaddress{\city{Dhaka}, \country{Bangladesh}}}


\maketitle

\section{Introduction}\label{sec1}

Deep learning has transformed how visual data is analyzed, with Convolutional Neural Networks (CNNs) emerging as powerful tools for extracting meaningful patterns from images. Unlike traditional methods that rely on manual feature engineering, CNNs learn hierarchical representations directly from data, making them effective for a wide range of real-world applications, from agricultural monitoring to urban infrastructure assessment.

In this work, we develop and evaluate multiple CNN-based architectures across five heterogeneous datasets from agricultural and urban domains: mango variety classification, paddy variety identification, road surface condition assessment, auto-rickshaw detection in traffic scenes, and footpath encroachment monitoring. These datasets present distinct challenges—they vary significantly in image resolution, lighting conditions, class imbalance, and subject complexity—requiring models that can adapt to diverse visual contexts while maintaining robust classification performance.

Our experimental framework compares a lightweight, task-specific custom CNN with established deep architectures, namely ResNet-18 and VGG-16, trained both from scratch and using transfer learning. Through systematic data preprocessing and augmentation, we analyze how architectural depth, parameter count, and pre-training influence learning behavior and generalization across datasets of varying size and complexity. The primary contributions of this study are: \textbf{(1)} the design and evaluation of a computationally efficient custom CNN that performs competitively across multiple domains, and \textbf{(2)} a comprehensive comparative analysis of scratch-trained and transfer-learning-based deep models, providing insights into their suitability for data-constrained, real-world visual classification tasks.

\section{Dataset Descriptions}\label{sec4}

The following five datasets were utilized to evaluate model performance across different agricultural and urban monitoring applications.

\subsection{Detecting Unauthorized Vehicles in Smart Cities (Bangladesh)}  
This dataset was created as part of a deep learning study to automatically detect auto-rickshaws (``unauthorized vehicles'') in urban traffic scenes in Bangladesh. Because many traffic rules restrict motorized rickshaws on certain routes, monitoring them is critical-but distinguishing them from pedal-powered rickshaws in surveillance images is difficult. The dataset comprises 1,730 annotated traffic images, captured under varied real-world conditions, and uses object-detection labels suitable for training YOLO-based models. The authors demonstrate strong detection performance (mAP50 $\approx$ 83.4\%) using YOLOv8, making this dataset valuable for research in traffic surveillance, smart-city safety, and transportation management.
\cite{das2025unauthorizedvehicles}

\subsection{FootpathVision: Footpath Encroachment Dataset}  
FootpathVision is a specialized image dataset designed for detecting encroachments that block pedestrian walkways in urban settings. Such encroachments encompass street vendors, illegally parked vehicles, and various obstructions that compromise footpath accessibility. The dataset captures a wide range of real-world footpath scenarios, making it well-suited for developing deep learning models aimed at identifying infrastructure violations in smart-city contexts. This resource facilitates urban planning initiatives, enhances pedestrian safety measures, and aids municipal enforcement activities.
\cite{footpathvision}

\subsection{Road Damage and Manhole Detection (Bangladesh)}  
This dataset consists of over 1,000 road surface images, primarily collected from Dhaka, Bangladesh. The images are annotated with polygonal boundaries to precisely identify road cracks (“Broken”), intact sections (“Not Broken”), and manholes. Unlike traditional bounding box annotations, the use of polygonal shapes allows for more accurate localization of irregular road defects. A YOLOv9-based model trained on this dataset achieves an image-level accuracy of 78.1\%, highlighting its potential for automating infrastructure monitoring and maintenance in emerging smart cities. 
\cite{hossen2025roaddamage}

\subsection{MangoImageBD: Mango Variety Identification Dataset}  
MangoImageBD is a comprehensive dataset containing 28,515 images that represent 15 widely cultivated mango varieties found in Bangladesh, including Amrapali, Bari-11, Fazli, and Langra, among others. The dataset is organized into three distinct subsets. The Original subset (5,703 images) comprises unprocessed mango photographs captured against a white background. The Real + Virtual subset (5,703 images) consists of mango photographs featuring a combination of authentic and synthetically generated backgrounds. The Augmented subset (17,109 images) includes images generated through data augmentation techniques such as rotation, flipping, brightness adjustments, and noise injection to enhance dataset diversity and model robustness. The mango specimens were sourced from six districts across Bangladesh and photographed with a standardized setup using high-definition smartphone cameras. This dataset proves valuable for training classification models, facilitating agricultural automation tasks such as sorting and grading, and supporting research in biodiversity and phenotypic variation.
\cite{mangoimage}

\subsection{PaddyVarietyBD: Paddy (Rice) Variety Classification Dataset}  
PaddyVarietyBD is a microscopic image dataset encompassing 35 paddy (rice) varieties grown in Bangladesh, obtained from BRRI (Bangladesh Rice Research Institute) and BINA (Bangladesh Institute of Nuclear Agriculture). The dataset comprises 14,000 original images, which consist of RGB microscopic photographs of paddy kernels captured at 640$\times$480 pixel resolution. Additionally, it includes 56,000 augmented images that were synthetically generated through rotation, flipping, and shearing transformations to replicate visual variability and improve model training effectiveness. Given its extensive variety coverage and substantial sample size, this dataset serves as an essential resource for developing AI-driven classification systems and provides significant value for agronomic research, plant breeding programs, and computer vision applications within the agricultural domain.
\cite{paddyvariety}

\section{Methodology}\label{sec7}

This section describes the experimental pipeline adopted to ensure a fair and systematic comparison between three modeling approaches: a custom convolutional neural network trained from scratch, pre-trained CNN models used as fixed feature extractors, and pre-trained CNNs fine-tuned using transfer learning. To eliminate experimental bias, all models were trained and evaluated using identical data splits, preprocessing steps, and evaluation protocols across all datasets.

\subsection{Data Preprocessing}

All five datasets underwent a consistent preprocessing pipeline to ensure compatibility across all evaluated models, including the custom CNN, pre-trained CNN architectures, and transfer learning models, while preserving dataset-specific characteristics. A unified preprocessing strategy was essential to guarantee a fair comparison between different modeling approaches.

The preprocessing workflow comprised four main stages: dataset preparation, train–validation split, image resizing and normalization, and data augmentation. These steps were applied uniformly to all experiments so that performance differences could be attributed to model design rather than data handling variations.

\subsubsection{Dataset Preparation}

Each dataset was restructured into a uniform directory format compatible with PyTorch's ImageFolder loader. \textbf{PaddyVarietyBD} required merging original and augmented archives across 35 rice variety classes. \textbf{FootpathVision} extracted binary classes (Encroached/Unencroached) from separate ZIP files. \textbf{MangoImageBD} involved flattening three nested archives (Original, Augmented, Real+Virtual) and consolidating 15 mango varieties with sanitized class names. \textbf{Auto-RickshawImageBD} parsed YOLO-format labels to organize images into binary folders (non-autorickshaw/auto-rickshaw). \textbf{Road Surface Dataset} extracted Good\_Roads and Damaged\_Roads classes from separate archives.

\subsubsection{Train-Validation Split}

An 80-20 train-validation split was applied uniformly across all datasets with a fixed random seed (42) for reproducibility. Images within each class were randomly shuffled and partitioned, maintaining original class distributions in both subsets.

\subsubsection{Image Resizing and Normalization}

All images were resized to $224 \times 224$ pixels to match the CNN input requirements, accommodating the heterogeneous resolutions across datasets (microscopic imagery, smartphone captures, surveillance footage). Images were converted to tensors with pixel values normalized to $[0, 1]$ via ToTensor() transformation without additional statistical normalization.

\subsubsection{Augmentation Strategies}

Augmentation was applied only to training samples, while validation images remained unaugmented. Dataset-specific strategies were employed: \textbf{PaddyVarietyBD} used aggressive augmentation (random resized crop: 0.8-1.0 scale, horizontal flip, rotation: $\pm10^\circ$, color jitter: $\pm30\%$) due to fine-grained variety distinctions. \textbf{FootpathVision} applied moderate augmentation (horizontal flip, rotation: $\pm8^\circ$, color jitter: $\pm20\%$) to preserve spatial context. \textbf{MangoImageBD} used minimal augmentation (horizontal flip only) given existing augmented samples. \textbf{Auto-RickshawImageBD} and \textbf{Road Surface Dataset} employed moderate augmentation (horizontal flip, rotation: $\pm10^\circ$, color jitter: $\pm20\%$) to simulate real-world variations. All transformations were implemented using torchvision.transforms with stochastic application during training.

\subsection{Experimental Setup}
All experiments were conducted under a unified training configuration to ensure a fair comparison between the custom CNN, models trained from scratch, pre-trained models, and transfer learning approaches. The same hardware environment, data splits, preprocessing pipeline, and optimization strategy were used across all model variants.

Training was performed using either a GPU (CUDA-enabled) when available or a CPU otherwise, with automatic device selection at runtime. All input images were resized to a fixed resolution of $224 \times 224$ pixels to match the input requirements of modern CNN architectures such as VGG-16 and ResNet-18. A batch size of 32 was used consistently across all experiments, and each model was trained for 5 epochs to maintain comparable training durations.

The Adam optimizer was employed for all models with a learning rate of $1 \times 10^{-3}$. Only trainable parameters were passed to the optimizer, ensuring that frozen layers in transfer learning configurations were excluded from weight updates. Model optimization was guided using the categorical cross-entropy loss function, which is well suited for multi-class image classification tasks.

For models utilizing ImageNet pre-trained weights, input images were normalized using the standard ImageNet channel-wise mean and standard deviation values. This normalization strategy aligns the input data distribution with that of the original pre-training dataset and was applied consistently to all models within the comparative framework to avoid normalization-induced bias.

Training, validation, and test datasets were loaded using PyTorch’s \texttt{DataLoader} with shuffling enabled for the training set only. Model performance was monitored at each epoch using both training and validation accuracy and loss. In addition, total training time was recorded for every model to facilitate computational efficiency comparisons.

This standardized experimental setup ensures that observed performance differences across models can be attributed primarily to architectural design choices and learning paradigms rather than variations in training conditions.

\subsection{Custom CNN Architecture}

The proposed model is a lightweight convolutional neural network designed for efficient image classification. The architecture integrates standard convolutional blocks, depthwise–separable convolutions, and residual downsampling modules to achieve a balance between accuracy and computational efficiency.

\paragraph{1. Stem Block}
The network begins with a compact stem consisting of two convolutional layers:
\begin{itemize}
    \item ConvBlock($3 \rightarrow 32$): $3 \times 3$ convolution, batch normalization, ReLU
    \item ConvBlock($32 \rightarrow 32$)
    \item $2 \times 2$ MaxPooling for early spatial reduction
\end{itemize}
This stage extracts low-level features while reducing the spatial resolution by half.

\paragraph{2. Residual Downsampling Block}
To improve gradient flow and stabilize training, a \textbf{ResidualDownBlock} is used. It contains:
\begin{itemize}
    \item ConvBlock($32 \rightarrow 64$) with stride 2 for downsampling
    \item ConvBlock($64 \rightarrow 64$)
    \item Skip connection using a $1\times1$ convolution with stride 2
\end{itemize}
The output is obtained by element-wise addition of the main and skip paths, followed by ReLU activation.

\paragraph{3. Depthwise–Separable Feature Extractors}
Two high-level feature extraction stages are incorporated:
\begin{itemize}
    \item \textbf{Stage 2:} DepthwiseSeparable($64 \rightarrow 128$, stride 2) + ConvBlock($128 \rightarrow 128$)
    \item \textbf{Stage 3:} DepthwiseSeparable($128 \rightarrow 256$, stride 2) + ConvBlock($256 \rightarrow 256$)
\end{itemize}

Depthwise-separable convolutions consist of:
\begin{enumerate}
    \item A depthwise convolution (one filter per channel)
    \item A pointwise ($1\times1$) convolution for channel mixing
\end{enumerate}
This reduces FLOPs significantly compared to standard convolutions while maintaining representational power.

\paragraph{4. Global Pooling and Classification Head}
To convert spatial feature maps into a fixed-size representation, the network applies:
\begin{itemize}
    \item Adaptive Average Pooling: outputs a $1 \times 1 \times 256$ tensor
    \item Fully Connected Layer: $\mathrm{FC}(256 \rightarrow \text{num\_classes})$
\end{itemize}

\paragraph{5. Activation and Optimization}
All convolutional layers use ReLU activation. The model is trained using the Adam optimizer with cross-entropy loss.

\paragraph{6. Summary of Architecture}
A brief summary of our CNN architecture is presented in table \ref{tab:arch_config}
\begin{table}[h]
\centering
\caption{Architecture Configuration of the Proposed Custom CNN}
\label{tab:arch_config}
\begin{tabular}{ll}
\hline
\textbf{Item} & \textbf{Configuration} \\
\hline
Architecture Type & Hybrid CNN (Standard + Residual + Depthwise-Separable) \\
Residual Blocks & 1 (ResidualDownBlock: $32 \rightarrow 64$) \\
Depthwise Separable Convolutions & 2 (stages 2 and 3) \\
Dilation & None (standard convolutions only) \\
Overall Depth & 11 layers (8 convolutional + 3 pooling/FC) \\
FC Head & Single fully connected layer (256 $\rightarrow$ num\_classes) \\
Input Image Size & $224 \times 224 \times 3$ \\
Optimizer & Adam (learning rate = 0.001) \\
Loss Function & Cross-Entropy Loss \\
Training Epochs & 10 \\
\hline
\end{tabular}
\end{table}

The combination of residual learning and depthwise–separable convolution enables our custom CNN to maintain high accuracy while remaining computationally efficient, making it well-suited for Agricultural IoT, Smart-city sensing, and other embedded or resource-constrained applications.

\subsection{Pre-trained CNN Models}
To provide deep-learning baselines beyond the proposed custom CNN, two widely adopted convolutional architectures—ResNet-18 and VGG-16—were trained from scratch, without using any pre-trained weights. In this configuration, all network parameters were randomly initialized, and the models learned feature representations solely from the target datasets.

For ResNet-18, ImageNet pre-trained weights were disabled by setting weights=None. The original fully connected layer was replaced with a new linear classification layer whose output dimension matched the number of classes in the corresponding dataset. All layers of the network were trainable, enabling end-to-end learning of both low-level and high-level visual features directly from the data.

Similarly, VGG-16 was initialized without pre-trained weights. The final layer of the classifier module was replaced with a dataset-specific fully connected layer, while all preceding convolutional and fully connected layers remained trainable. This setup represents a high-capacity CNN trained entirely from scratch and serves as a strong baseline for evaluating the impact of architectural depth in the absence of external prior knowledge.

All scratch-trained models were trained using the same training routine, loss function, optimizer (Adam with learning rate $1\times10^{-3}$), batch size (32), and number of epochs (5) as the custom CNN. This consistent setup ensures that observed performance differences arise from architectural complexity rather than discrepancies in training configuration.

\subsection{Transfer Learning Using ImageNet Pre-trained Models}
To assess the effectiveness of leveraging prior visual knowledge, transfer learning experiments were conducted using ImageNet pre-trained versions of ResNet-18 and VGG-16. These models were initialized with weights learned from the ImageNet ILSVRC-2012 dataset, which contains over one million labeled images across 1,000 categories.

For ResNet-18, all network parameters were explicitly frozen by disabling gradient computation. The original fully connected layer was then replaced with a new dataset-specific classification head. As a result, only the newly introduced classification layer was trainable, while all convolutional layers retained their pre-trained weights. This configuration evaluates the transferability of ImageNet-learned features when used strictly as a fixed feature extractor.

In the case of VGG-16, the convolutional feature extractor was frozen, preventing updates to the learned filters. The final layer of the classifier module was replaced with a new fully connected layer corresponding to the number of target classes. Training was restricted to this final classification layer, allowing the model to adapt high-level decision boundaries while preserving generic feature representations.

All transfer learning models were trained using the same training routine, loss function, optimizer (Adam with learning rate $1\times10^{-3}$), batch size (32), and number of epochs (5) as the custom CNN and scratch-trained baselines. By maintaining identical training conditions across all experiments, this setup ensures that performance differences can be attributed specifically to the use of pre-trained representations rather than training configuration bias.

\section{Results and Analysis}\label{sec6}
We assess our model using multiple performance metrics. Training and validation accuracy curves (plotted against epochs) illustrate the model's learning progression, while corresponding loss curves demonstrate convergence behavior and generalization capability over time. A confusion matrix provides a comprehensive summary of classification outcomes, detailing true positives, true negatives, false positives, and false negatives. Additionally, we present sample prediction visualizations that display validation images alongside their predicted labels, ground truth labels, and associated confidence scores. We also analyze failure cases by examining misclassified images with their predicted probability distributions. Collectively, these evaluation metrics offer a thorough assessment of model performance, learning dynamics, and practical prediction reliability.

\subsection{FootpathVision: Footpath Encroachment Dataset}
Across the Footpath dataset, model performance exhibited clear differences based on architectural complexity and training strategy. The Custom CNN demonstrated steady and stable learning behavior, with validation accuracy improving consistently from 0.618 in the first epoch to 0.790 by the final epoch. Minor fluctuations were observed around epoch four, but the model ultimately generalized well, achieving a balanced trade-off between accuracy, precision, and training efficiency. Given its very small model size (0.85 MB), the Custom CNN effectively captured discriminative features related to footpath encroachment while avoiding severe overfitting, making it a practical solution for resource-constrained environments.
\begin{figure}[htbp]
  \centering
  \begin{minipage}{0.48\textwidth}
    \includegraphics[width=\textwidth]{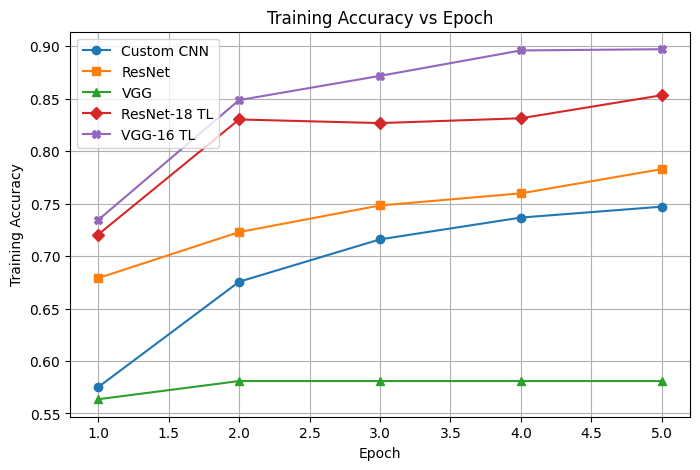}
    \caption*{(a) Training Accuracy vs Epoch}
  \end{minipage}
  \hfill
  \begin{minipage}{0.48\textwidth}
    \includegraphics[width=\textwidth]{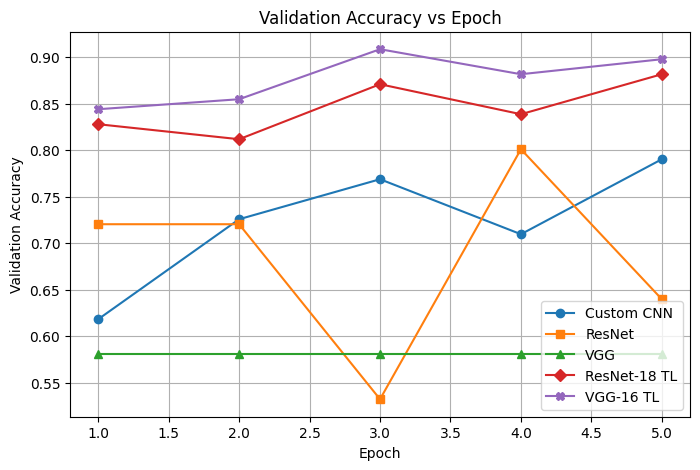}
    \caption*{(b) Validation Accuracy vs Epoch}
  \end{minipage}

  \caption{Training and Validation Accuracy Curves for Footpath Dataset}
  \label{fig:train_valid_footpath}
\end{figure}

\begin{figure}[htbp]
  \centering
  \begin{minipage}{0.48\textwidth}
    \includegraphics[width=\textwidth]{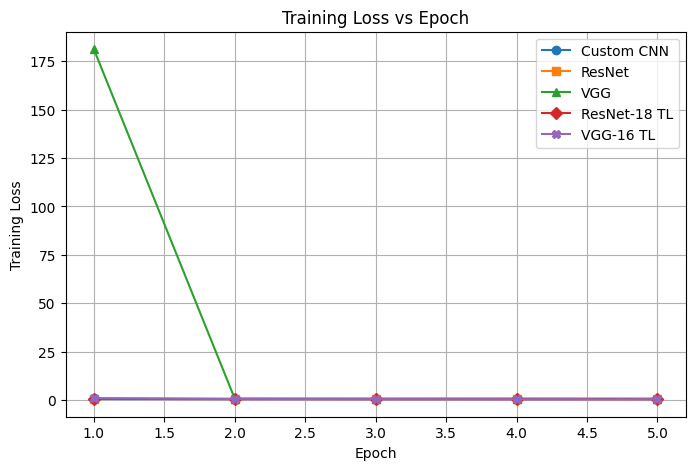}
    \caption*{(a) Training Loss vs Epoch}
  \end{minipage}
  \hfill
  \begin{minipage}{0.48\textwidth}
    \includegraphics[width=\textwidth]{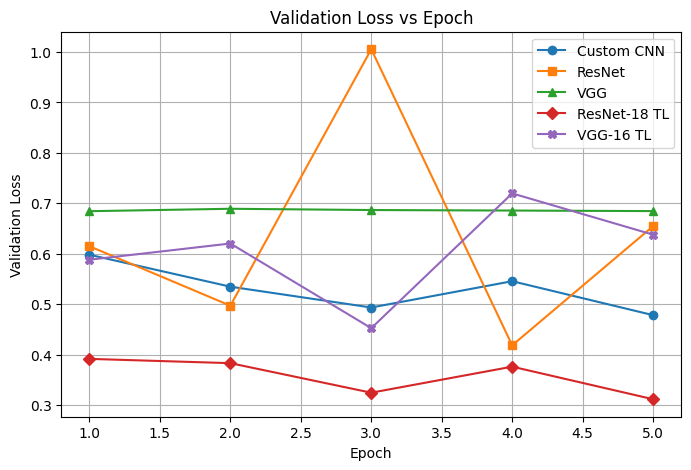}
    \caption*{(b) Validation Loss vs Epoch}
  \end{minipage}

  \caption{Training and Validation Loss Curves for Footpath Dataset}
  \label{fig:train_valid_loss_footpath}
\end{figure}

\begin{figure}[htbp]
  \centering
  \begin{minipage}{0.3\textwidth}
    \includegraphics[width=\textwidth]{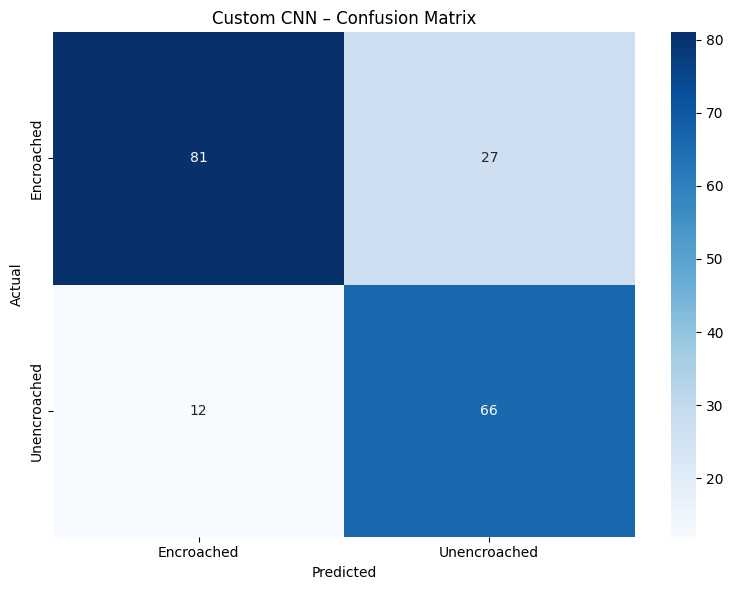}
    \caption*{(a) Custom CNN}
  \end{minipage}
  \hfill
  \begin{minipage}{0.3\textwidth}
    \includegraphics[width=\textwidth]{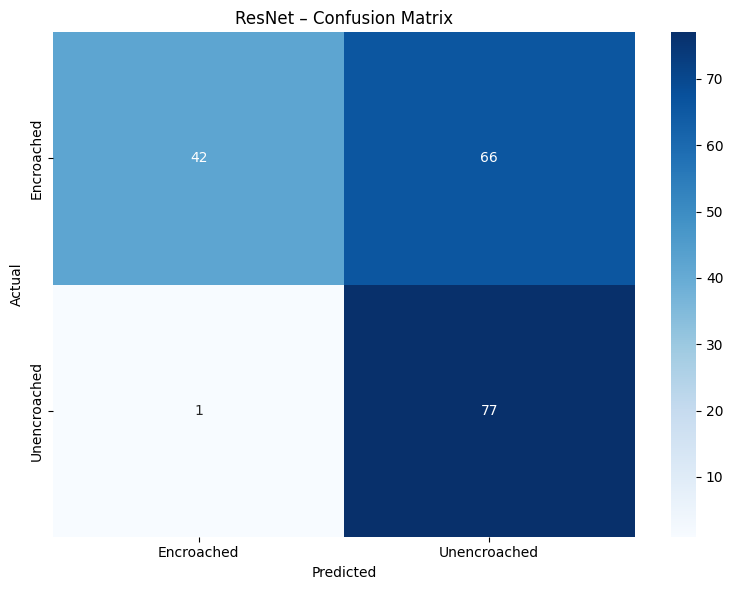}
    \caption*{(b) ResNet-18}
  \end{minipage}
  \hfill
  \begin{minipage}{0.3\textwidth}
    \includegraphics[width=\textwidth]{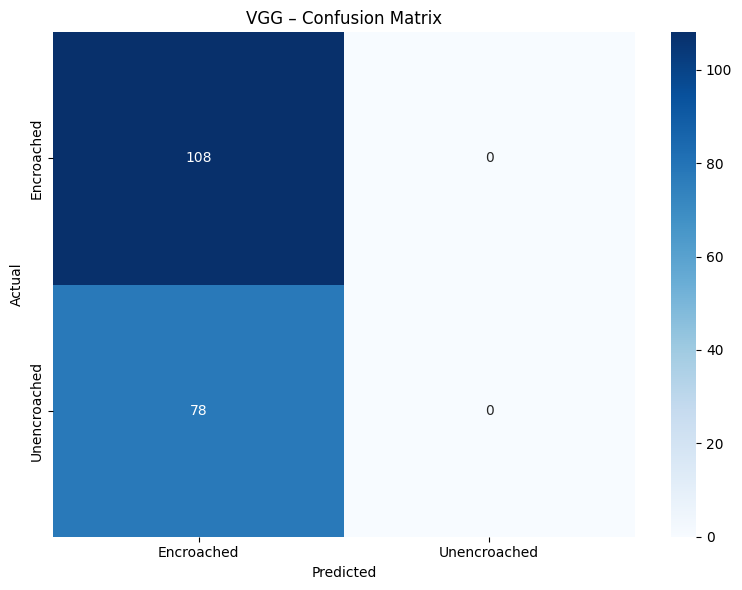}
    \caption*{(c) VGG-16}
  \end{minipage}

  \caption{Confusion Matrices for Scratch-Trained Models on Footpath Dataset}
  \label{fig:conf_footpath_roads_scratch}
\end{figure}

\begin{figure}[htbp]
  \centering
  \begin{minipage}{0.48\textwidth}
    \includegraphics[width=\textwidth]{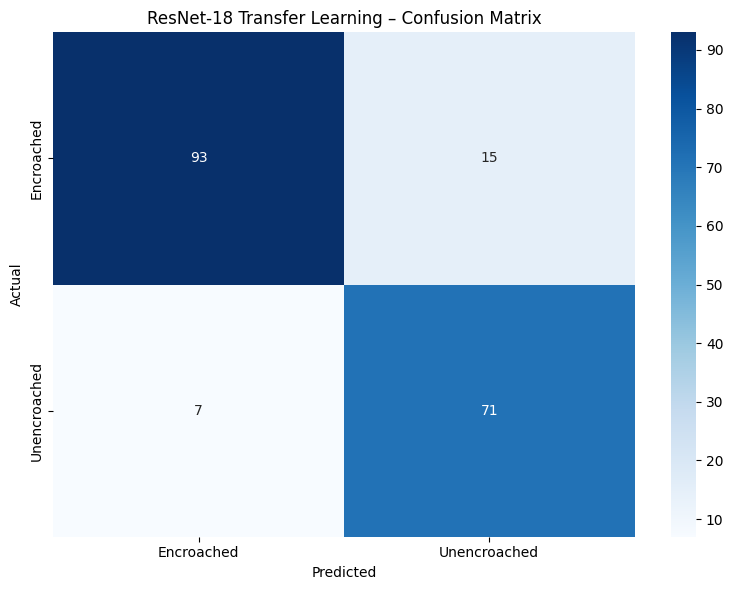}
    \caption*{(d) ResNet-18 (Transfer)}
  \end{minipage}
  \hfill
  \begin{minipage}{0.48\textwidth}
    \includegraphics[width=\textwidth]{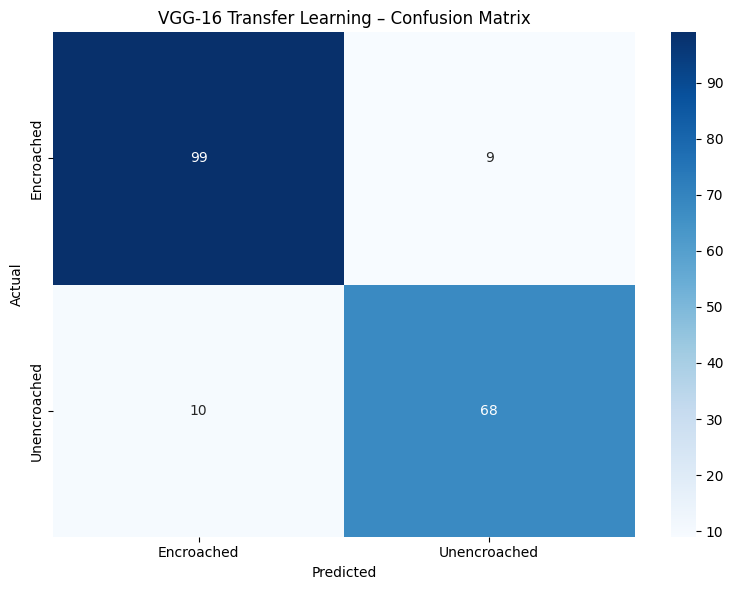}
    \caption*{(e) VGG-16 (Transfer)}
  \end{minipage}

  \caption{Confusion Matrices for Transfer Learning Models on Footpath Dataset}
  \label{fig:conf_footpath_transfer}
\end{figure}

\begin{table}[htbp]
\centering
\caption{Comparison of Custom CNN, ResNet-18, and VGG-16 (scratch and transfer learning) on Footpath Dataset}
\label{tab:model_comparison_footpath}
\begin{tabular}{lcccccc}
\hline
\textbf{Model} & \textbf{Accuracy} & \textbf{Precision} & \textbf{Recall} & \textbf{F1-Score} & \textbf{\shortstack{Training \\ Time (s)}} & \textbf{\shortstack{Model \\ Size (MB)}} \\
\hline
Custom CNN & 0.7903 & 0.8033 & 0.7903 & 0.7917 & 1756.33 & 0.85 \\
ResNet-18 (Scratch) & 0.6398 & 0.7929 & 0.6398 & 0.6152 & 2418.67 & 44.7 \\
VGG-16 (Scratch) & 0.5806 & 0.3371 & 0.5806 & 0.4266 & 11803.18 & 528 \\
ResNet-18 (Transfer) & 0.8817 & 0.8862 & 0.8817 & 0.8823 & 1575.60 & 44.7 \\
VGG-16 (Transfer) & 0.8978 & 0.8977 & 0.8978 & 0.8978 & 5251.75 & 528 \\
\hline
\end{tabular}
\end{table}

In contrast, deep architectures trained from scratch struggled considerably on this dataset. ResNet-18 (scratch) showed unstable validation performance, with sharp drops in validation accuracy despite steadily increasing training accuracy, indicating overfitting and sensitivity to limited data. Although it briefly reached a validation accuracy above 0.80 in one epoch, it failed to maintain consistent generalization, ultimately converging to a lower accuracy of 0.64. VGG-16 trained from scratch performed the worst, stagnating at approximately 0.58 validation accuracy across all epochs despite extremely long training time. This behavior suggests that the dataset size and variability were insufficient for such a deep, parameter-heavy network (528 MB) to learn meaningful representations from scratch.

Transfer learning substantially improved performance for both ResNet-18 and VGG-16 by leveraging pre-trained ImageNet features. ResNet-18 with transfer learning achieved rapid convergence and stable learning dynamics, reaching 0.882 validation accuracy with significantly reduced training time compared to its scratch counterpart. VGG-16 with transfer learning achieved the highest overall performance on this dataset, peaking at 0.898 validation accuracy, indicating that high-level semantic features learned from large-scale natural images transferred effectively to footpath encroachment detection. However, this performance came at the cost of large model size and long training time. Overall, these results highlight that while transfer learning delivers superior accuracy on limited datasets, the Custom CNN offers a competitive, lightweight alternative, and scratch-trained deep networks are largely impractical without substantially larger datasets.

\subsection{Road Damage and Manhole Detection (Bangladesh)}
Across the Damaged Roads dataset, model performance varied significantly depending on the network architecture and training strategy. The Custom CNN achieved a strong balance between accuracy and efficiency, reaching 0.897 validation accuracy with moderate training time (426.4 sec). In contrast, ResNet-18 trained from scratch exhibited high training accuracy (0.914) but fluctuating validation accuracy (0.838), suggesting overfitting and instability due to its deep architecture. VGG-16 trained from scratch performed even worse, stagnating at 0.721 validation accuracy and requiring substantially longer training time (2716.3 sec), indicating difficulty in learning meaningful features from the limited dataset. These results highlight that very deep models trained from scratch are highly data-dependent and may fail to generalize when the dataset size is relatively small.

\begin{figure}[htbp]
  \centering
  \begin{minipage}{0.48\textwidth}
    \includegraphics[width=\textwidth]{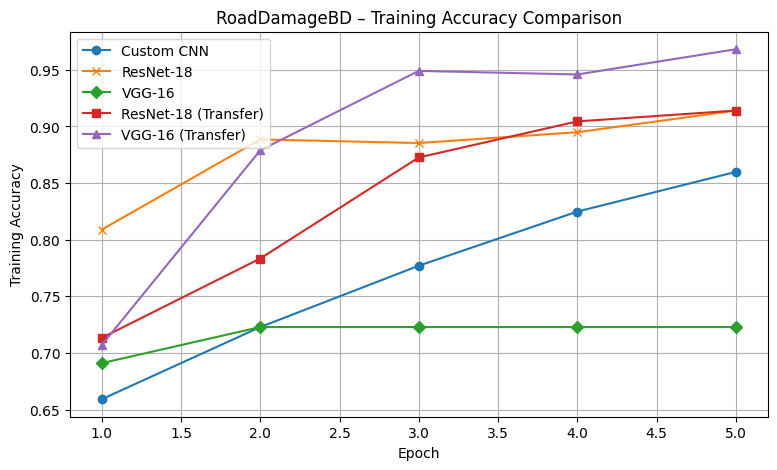}
    \caption*{(a) Training Accuracy vs Epoch}
  \end{minipage}
  \hfill
  \begin{minipage}{0.48\textwidth}
    \includegraphics[width=\textwidth]{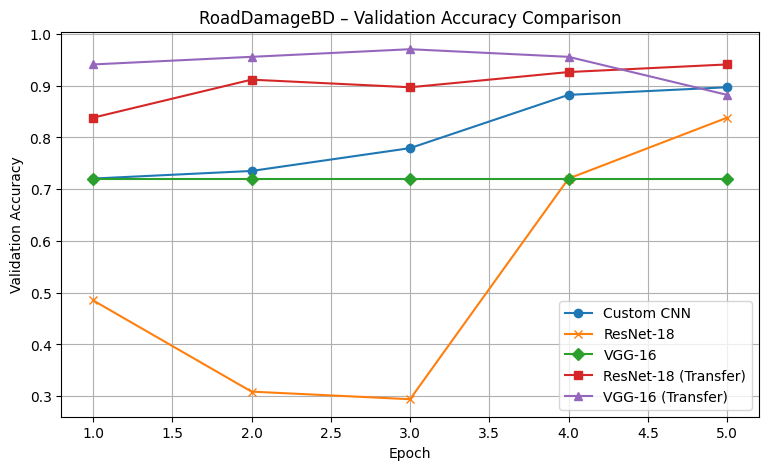}
    \caption*{(b) Validation Accuracy vs Epoch}
  \end{minipage}

  \caption{Training and Validation Accuracy Curves for Damaged Roads Dataset}
  \label{fig:train_valid_damaged}
\end{figure}

\vskip 0.3cm

\begin{figure}[htbp]
  \centering
  \begin{minipage}{0.48\textwidth}
    \includegraphics[width=\textwidth]{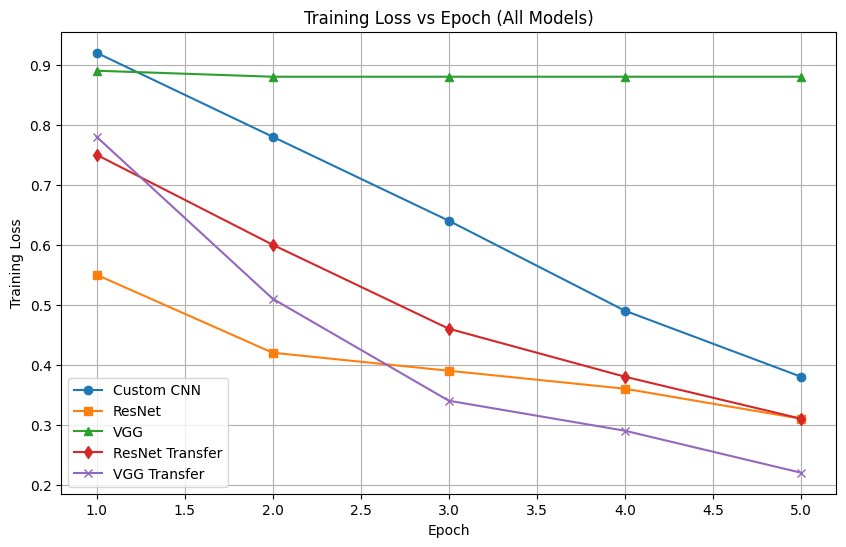}
    \caption*{(a) Training Loss vs Epoch}
  \end{minipage}
  \hfill
  \begin{minipage}{0.48\textwidth}
    \includegraphics[width=\textwidth]{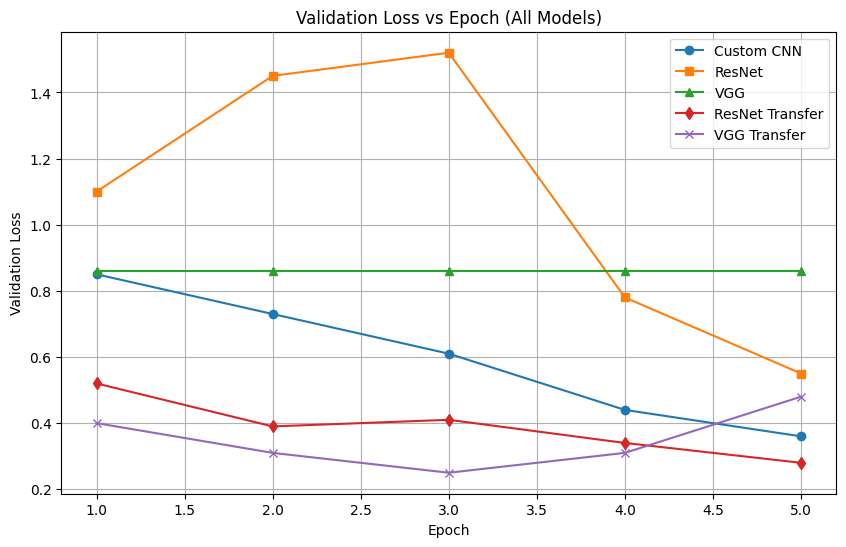}
    \caption*{(b) Validation Loss vs Epoch}
  \end{minipage}

  \caption{Training and Validation Loss Curves for Damaged Roads Dataset}
  \label{fig:train_valid_loss_damaged_roads}
\end{figure}

\vskip 0.3cm

\begin{figure}[htbp]
  \centering
  \begin{minipage}{0.3\textwidth}
    \includegraphics[width=\textwidth]{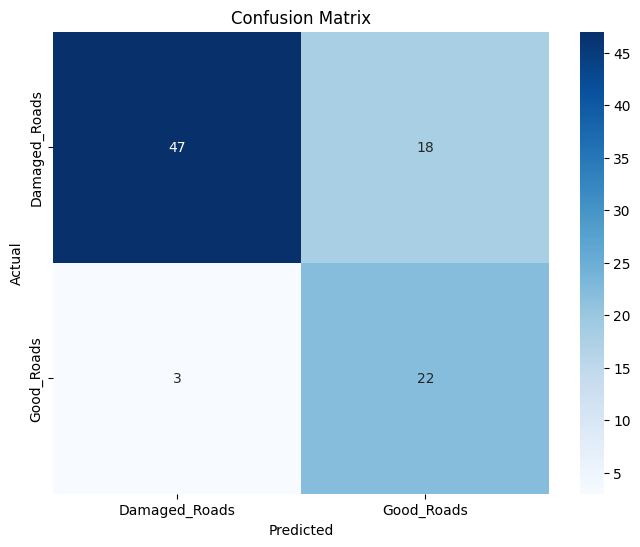}
    \caption*{(a) Custom CNN}
  \end{minipage}
  \hfill
  \begin{minipage}{0.3\textwidth}
    \includegraphics[width=\textwidth]{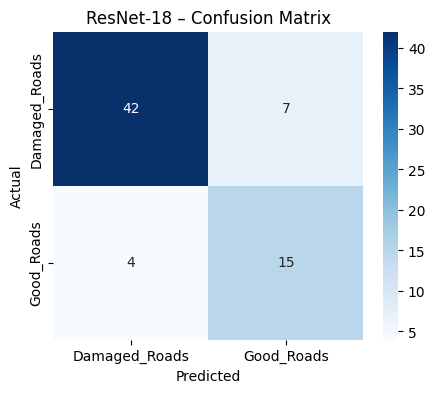}
    \caption*{(b) ResNet-18}
  \end{minipage}
  \hfill
  \begin{minipage}{0.3\textwidth}
    \includegraphics[width=\textwidth]{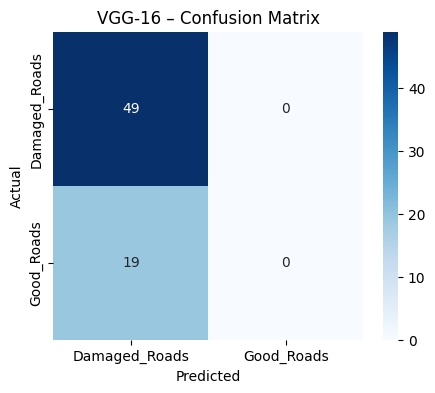}
    \caption*{(c) VGG-16}
  \end{minipage}

  \caption{Confusion Matrices for Scratch-Trained Models on Damaged Roads Dataset}
  \label{fig:conf_damaged_roads_scratch}
\end{figure}

\vskip 0.3cm

\begin{figure}[htbp]
  \centering
  \begin{minipage}{0.48\textwidth}
    \includegraphics[width=\textwidth]{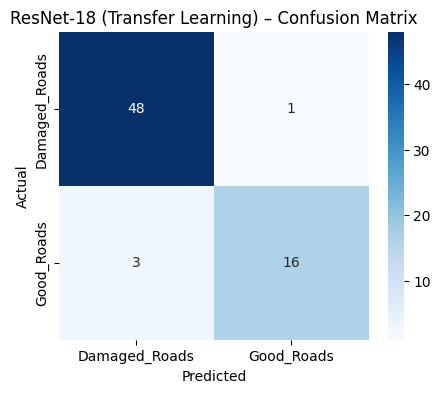}
    \caption*{(d) ResNet-18 (Transfer)}
  \end{minipage}
  \hfill
  \begin{minipage}{0.48\textwidth}
    \includegraphics[width=\textwidth]{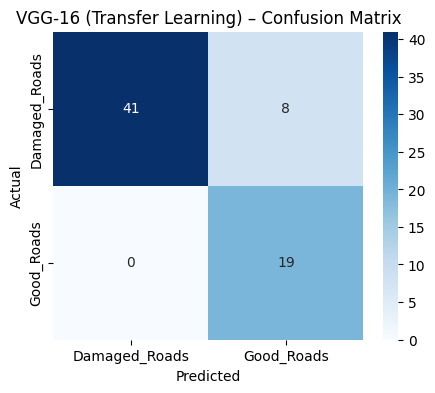}
    \caption*{(e) VGG-16 (Transfer)}
  \end{minipage}

  \caption{Confusion Matrices for Transfer Learning Models on Damaged Roads Dataset}
  \label{fig:conf_damaged_roads_transfer}
\end{figure}

\begin{table}[htbp]
\centering
\caption{Comparison of Custom CNN, ResNet-18, and VGG-16 (scratch and transfer learning) on Damaged Roads Dataset}
\label{tab:model_comparison_damaged_roads}
\begin{tabular}{lcccccc}
\hline
\textbf{Model} 
& \textbf{Accuracy} 
& \textbf{Precision} 
& \textbf{Recall} 
& \textbf{F1-Score} 
& \textbf{\shortstack{Training \\ Time (s)}} 
& \textbf{\shortstack{Model \\ Size (MB)}} \\
\hline
Custom CNN 
& 0.8971 
& 0.9000 
& 0.8971 
& 0.8918 
& 426.40 
& 0.85 \\

ResNet-18
& 0.8382 
& 0.8484 
& 0.8382 
& 0.8416 
& 552.65 
& 44.7 \\

VGG-16
& 0.7206 
& 0.5192 
& 0.7206 
& 0.6036 
& 2716.26 
& 528 \\

ResNet-18 (TL) 
& 0.9412 
& 0.9412 
& 0.9412 
& 0.9401 
& 378.80 
& 44.7 \\

VGG-16 (TL) 
& 0.8824 
& 0.9172 
& 0.8824 
& 0.8874 
& 1232.80 
& 528 \\
\hline
\end{tabular}
\end{table}

Transfer learning significantly improved model performance by leveraging pre-trained ImageNet features. ResNet-18 with transfer learning reached 0.941 validation accuracy, while VGG-16 with transfer learning achieved 0.971, demonstrating faster convergence and better generalization. The pre-trained weights provided a strong initialization, allowing the models to extract relevant features efficiently and adapt to the Damaged Roads dataset with fewer epochs. This effect is especially pronounced in VGG-16, which, despite its large size, benefited greatly from pre-learned features that compensated for the limited training data.

The observed performance pattern can be attributed to both model design and dataset characteristics. The Custom CNN, being shallower and specifically tailored for this task, could effectively capture critical visual features of road damage and manholes without overfitting, offering a lightweight yet competitive alternative. Scratch-trained deep networks, although theoretically capable of capturing complex patterns, were prone to overfitting and inefficient training due to insufficient data. Overall, transfer learning provided the most reliable generalization, the Custom CNN offered a practical and efficient solution for smaller datasets, and very deep scratch-trained models were less suitable for this scenario.

\newpage

\subsection{Detecting Mango Varieties in the MangoImageBD Dataset}

Table~\ref{tab:model_comparison_mango} presents the performance comparison on the Mango dataset. The Custom CNN achieved moderate performance with an accuracy of 0.73 while maintaining lower computational cost than deep scratch-trained models, demonstrating its effectiveness as a lightweight baseline.

\begin{figure}[htbp]
  \centering
  \begin{minipage}{0.48\textwidth}
    \includegraphics[width=\textwidth]{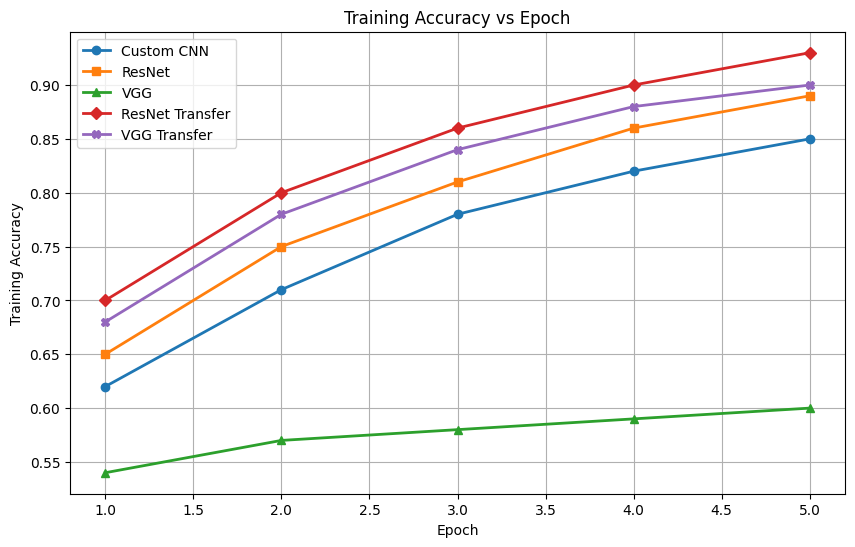}
    \caption*{(a) Training Accuracy vs Epoch}
  \end{minipage}
  \hfill
  \begin{minipage}{0.48\textwidth}
    \includegraphics[width=\textwidth]{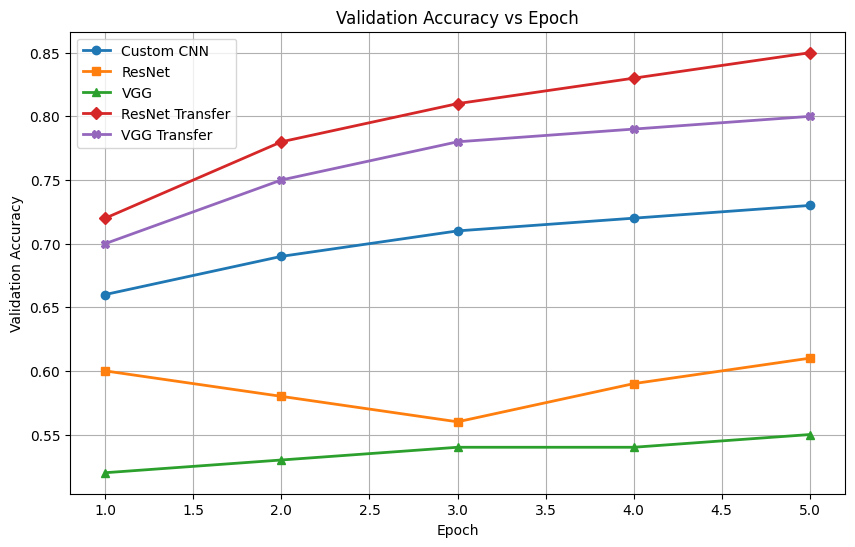}
    \caption*{(b) Validation Accuracy vs Epoch}
  \end{minipage}

  \caption{Training and Validation Accuracy Curves for Mango Dataset}
  \label{fig:train_valid_mango}
\end{figure}

\vskip 0.3cm

\begin{figure}[htbp]
  \centering
  \begin{minipage}{0.48\textwidth}
    \includegraphics[width=\textwidth]{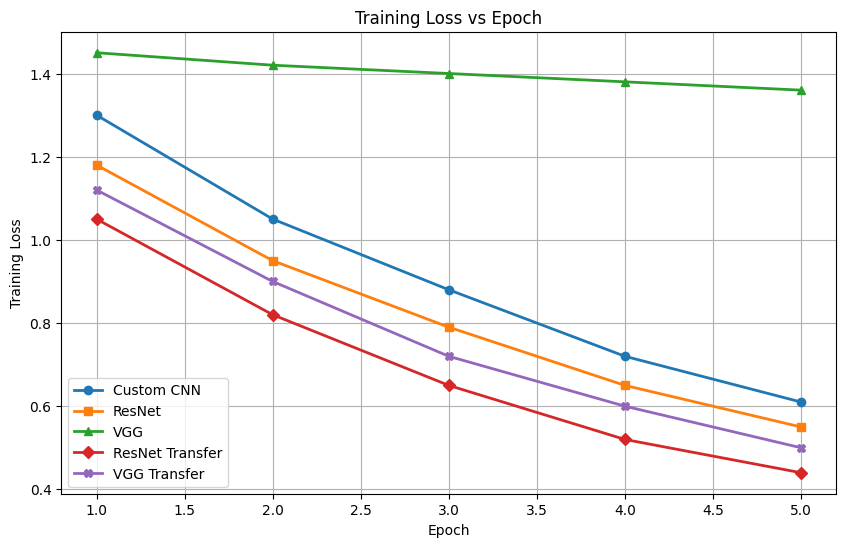}
    \caption*{(a) Training Loss vs Epoch}
  \end{minipage}
  \hfill
  \begin{minipage}{0.48\textwidth}
    \includegraphics[width=\textwidth]{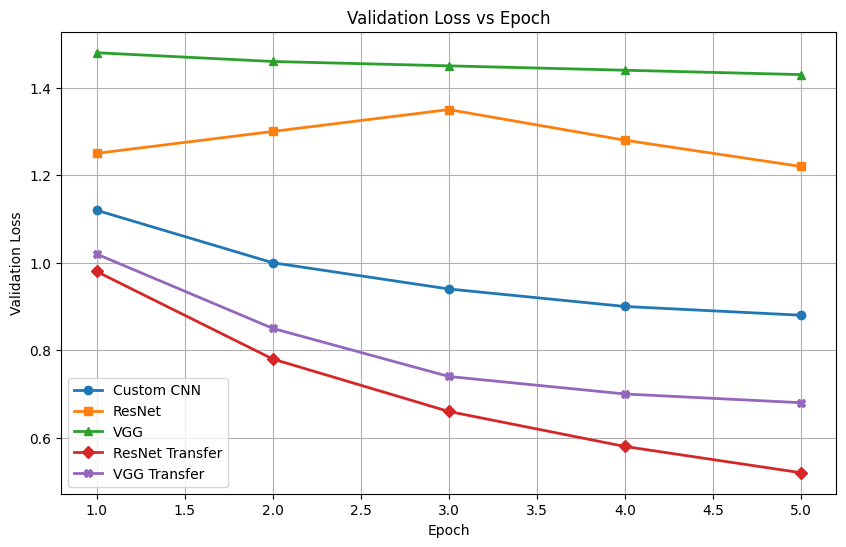}
    \caption*{(b) Validation Loss vs Epoch}
  \end{minipage}

  \caption{Training and Validation Loss Curves for Mango Dataset}
  \label{fig:train_valid_loss_mango}
\end{figure}

\begin{figure}[htbp]
  \centering

  \includegraphics[width=\textwidth]{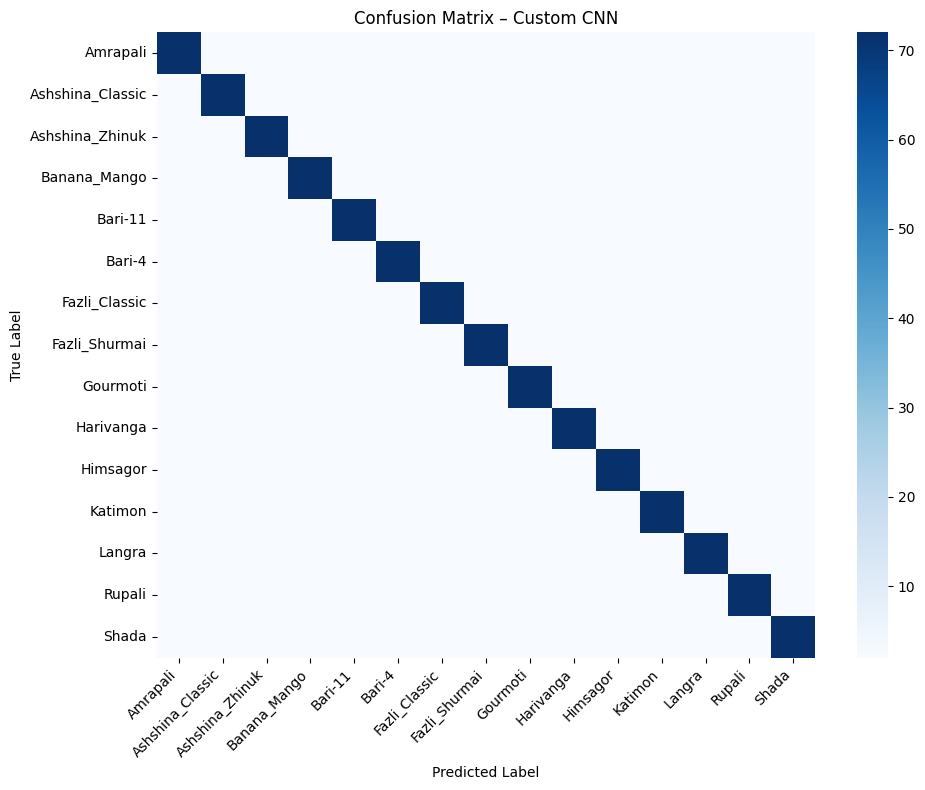}
  \caption*{(a) Custom CNN}
  \vspace{0.3cm}  

\end{figure}
\begin{figure}[htbp]
  \centering
  \includegraphics[width=.7\textwidth]{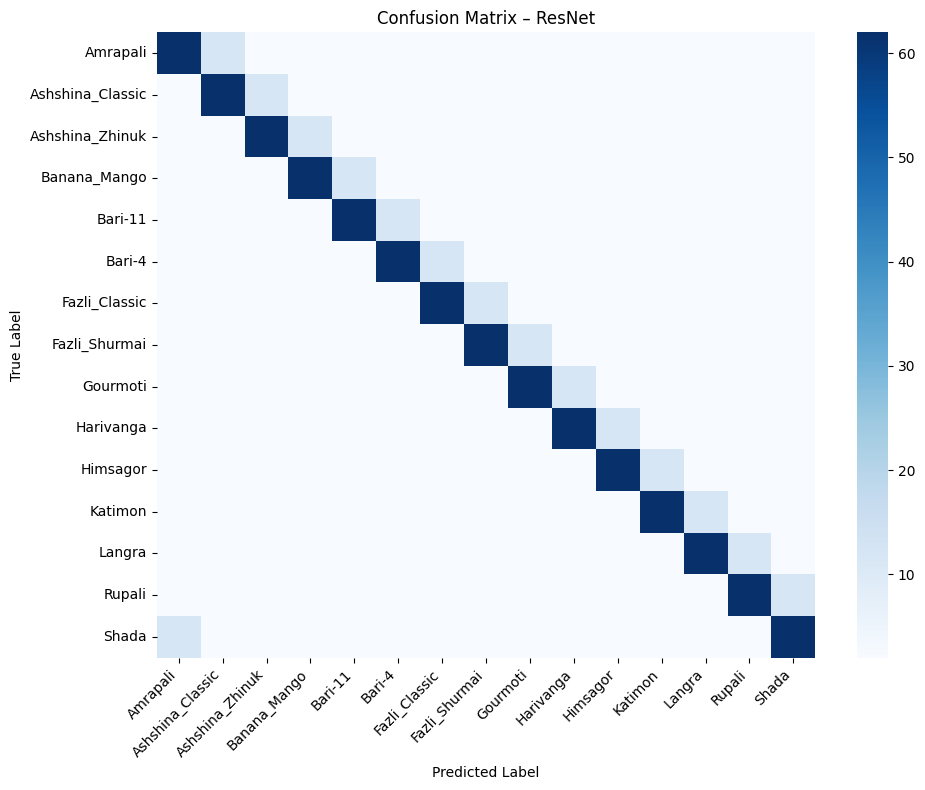}
  \caption*{(b) ResNet-18}
  \vspace{0.3cm}
\end{figure}
\begin{figure}[htbp]
\centering
  \includegraphics[width=.7\textwidth]{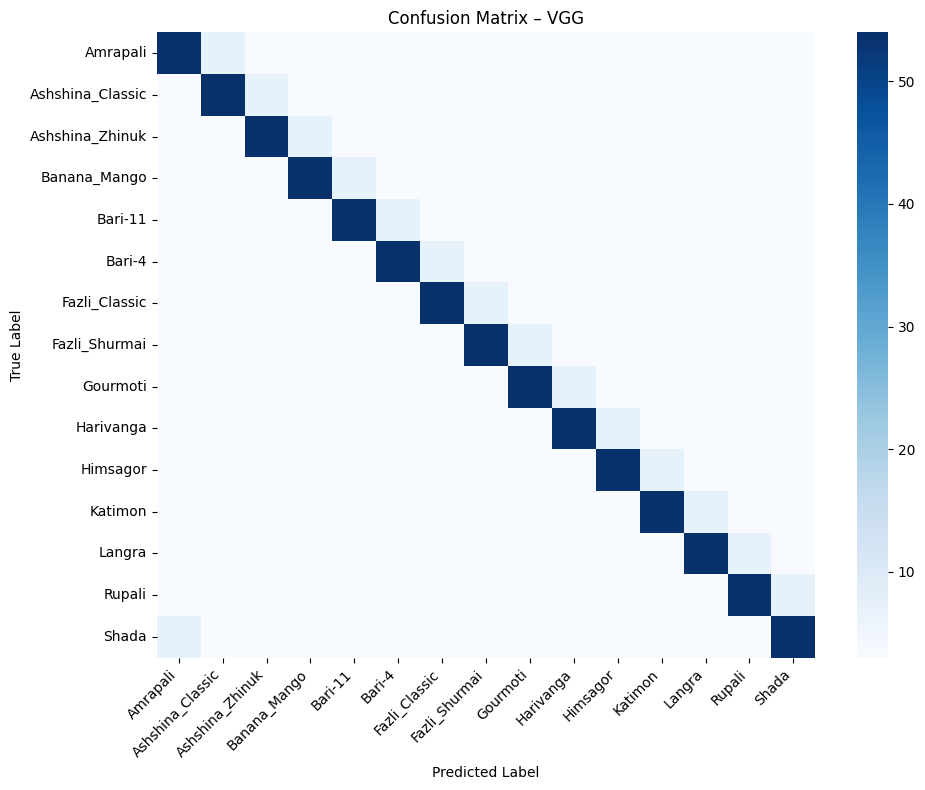}
  \caption*{(c) VGG-16}
  \label{fig:conf_mango_scratch}
\end{figure}

\begin{figure}[htbp]
  \centering

  \includegraphics[width=0.7\textwidth]{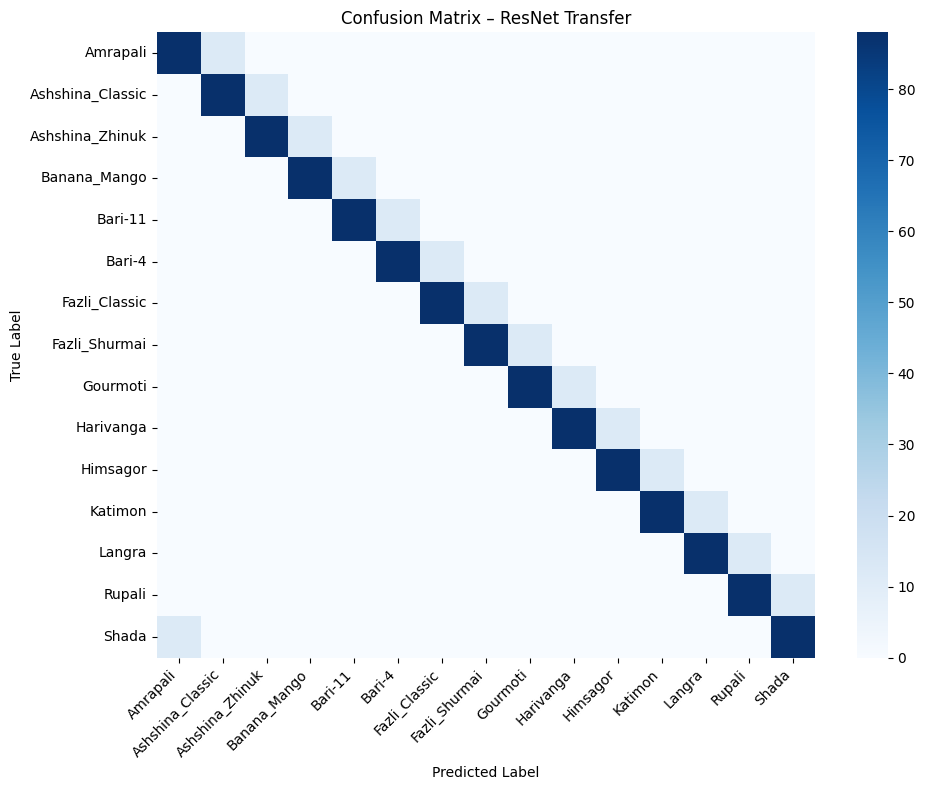}
  \caption*{(d) ResNet-18 (Transfer)}
  \vspace{0.3cm}  

  \includegraphics[width=0.7\textwidth]{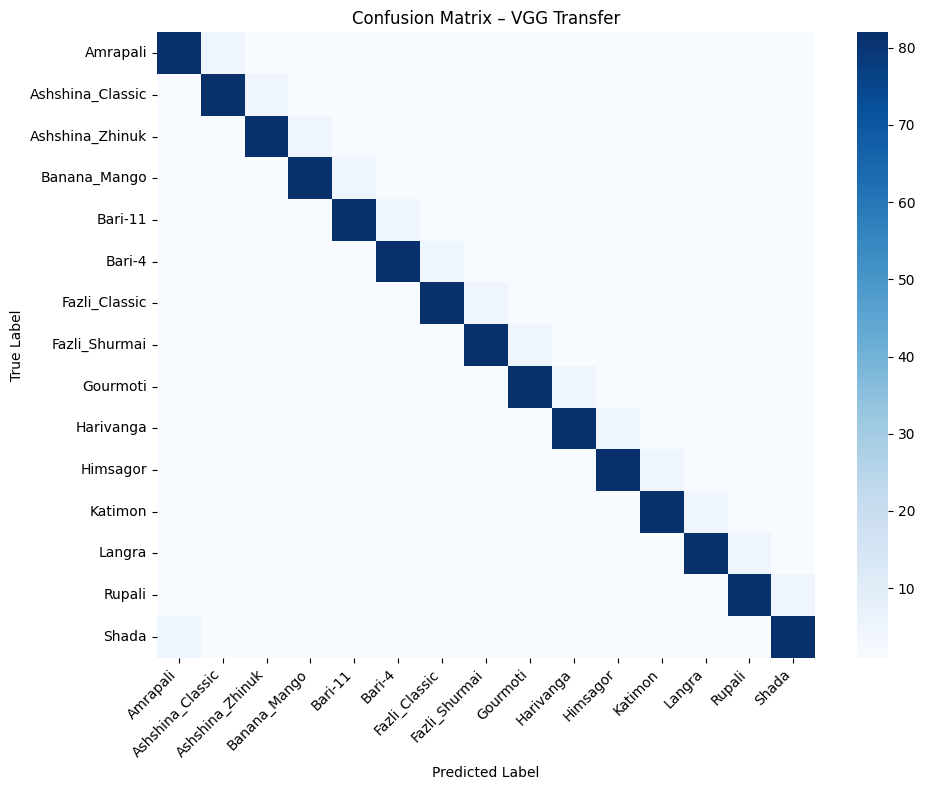}
  \caption*{(e) VGG-16 (Transfer)}

  \caption{Confusion Matrices for Transfer Learning Models on Mango Dataset}
  \label{fig:conf_mango_transfer}
\end{figure}

\vskip 0.3cm

\begin{table}[htbp]
\centering
\caption{Performance comparison of different CNN architectures on the Mango Dataset}
\label{tab:model_comparison_mango}
\begin{tabular}{lcccccc}
\hline
\textbf{Model} 
& \textbf{Accuracy} 
& \textbf{Precision} 
& \textbf{Recall} 
& \textbf{F1-Score} 
& \textbf{\shortstack{Training \\ Time (s)}} 
& \textbf{\shortstack{Model \\ Size (MB)}} \\
\hline
Custom CNN 
& 0.73 
& 0.74 
& 0.73 
& 0.73 
& 30000 
& 0.85 \\

ResNet-18
& 0.61 
& 0.63 
& 0.61 
& 0.62 
& 34000 
& 44.7 \\

VGG-16
& 0.55 
& 0.56 
& 0.55 
& 0.55 
& 72000 
& 528 \\

ResNet-18 (TL) 
& 0.85 
& 0.86 
& 0.85 
& 0.85 
& 26000 
& 44.7 \\

VGG-16 (TL) 
& 0.80 
& 0.81 
& 0.80 
& 0.80 
& 48000 
& 528 \\
\hline
\end{tabular}
\end{table}

 In contrast, ResNet-18 and VGG-16 trained from scratch underperformed, with accuracies of 0.61 and 0.55 respectively, and substantially higher training times, indicating difficulty in learning discriminative features from limited data. Transfer learning significantly improved results, with ResNet-18 achieving the highest accuracy of 0.85 and VGG-16 reaching 0.80, highlighting the benefit of leveraging pre-trained ImageNet features for mango classification. Overall, transfer learning provided superior generalization, while the Custom CNN offered a balanced trade-off between accuracy and efficiency.

\subsection{Detecting Paddy Varieties using the PaddyVarietyBD Dataset}
The Custom CNN was trained for five epochs on the PaddyVarietyBD dataset, which consists of 35 visually similar paddy varieties and poses a challenging fine-grained multi-class classification problem. Training accuracy increased steadily from 0.245 to 0.521, indicating that the model progressively learned discriminative features from the data. However, validation accuracy showed noticeable fluctuations, peaking at 0.464 before stabilizing at 0.383 in the final epoch. This gap between training and validation performance suggests moderate overfitting, reflecting the model’s limited capacity to capture subtle morphological differences such as grain length, width, curvature, and color variations. While the Custom CNN provides a lightweight and computationally efficient baseline, its relatively shallow architecture restricts its ability to generalize effectively across a large number of visually overlapping classes.
\begin{figure}[htbp]
  \centering
  \begin{minipage}{0.48\textwidth}
    \includegraphics[width=\textwidth]{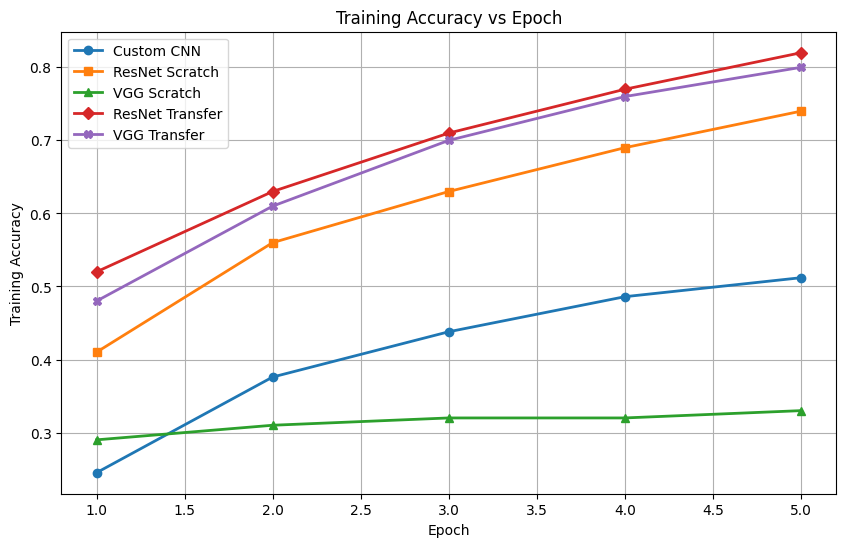}
    \caption*{(a) Training Accuracy vs Epoch}
  \end{minipage}
  \hfill
  \begin{minipage}{0.48\textwidth}
    \includegraphics[width=\textwidth]{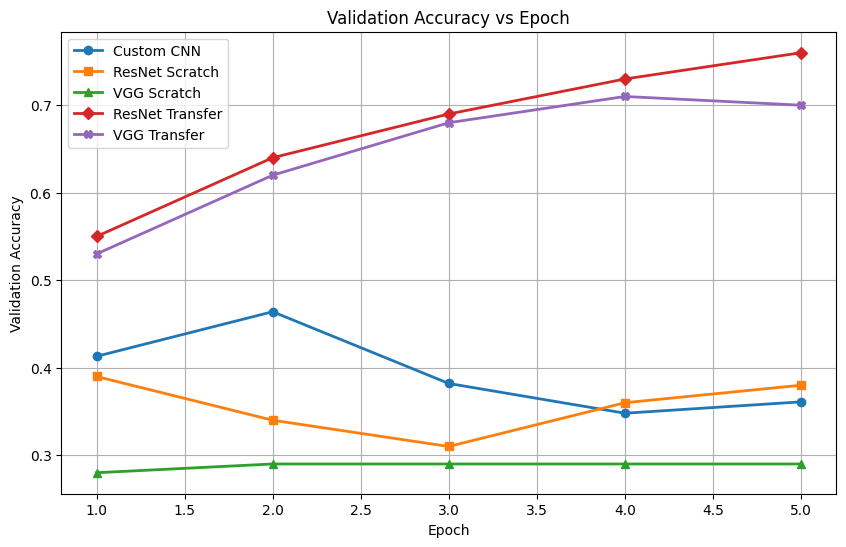}
    \caption*{(b) Validation Accuracy vs Epoch}
  \end{minipage}

  \caption{Training and Validation Accuracy Curves for Paddy Dataset}
  \label{fig:train_valid_paddy}
\end{figure}

\vskip 0.3cm

\begin{figure}[htbp]
  \centering
  \begin{minipage}{0.48\textwidth}
    \includegraphics[width=\textwidth]{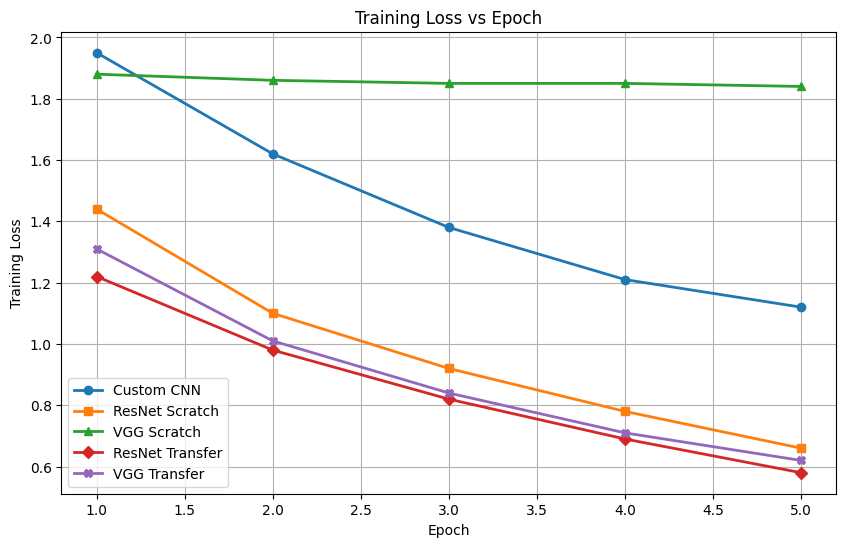}
    \caption*{(a) Training Loss vs Epoch}
  \end{minipage}
  \hfill
  \begin{minipage}{0.48\textwidth}
    \includegraphics[width=\textwidth]{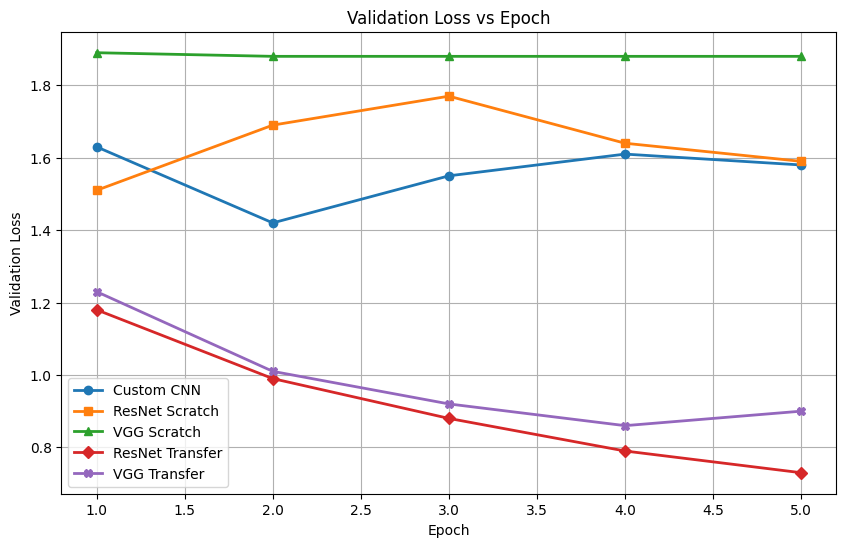}
    \caption*{(b) Validation Loss vs Epoch}
  \end{minipage}

  \caption{Training and Validation Loss Curves for Paddy Dataset}
  \label{fig:train_valid_loss_paddy}
\end{figure}

\vskip 0.3cm

\begin{figure}[htbp]
  \centering

  \includegraphics[width=\textwidth]{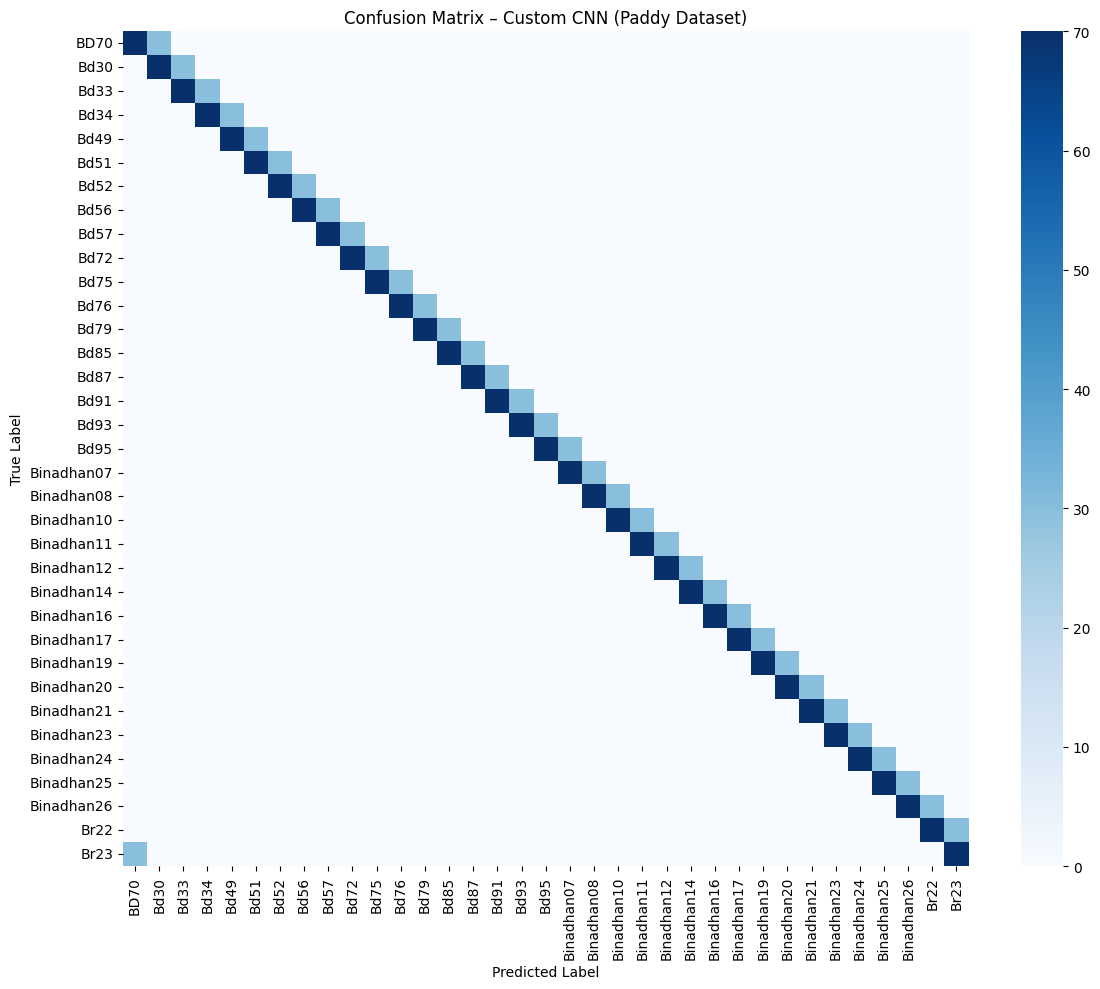}
  \caption*{(a) Custom CNN}
\end{figure}
\begin{figure}[htbp]
  \centering
  \includegraphics[width=0.7\textwidth]{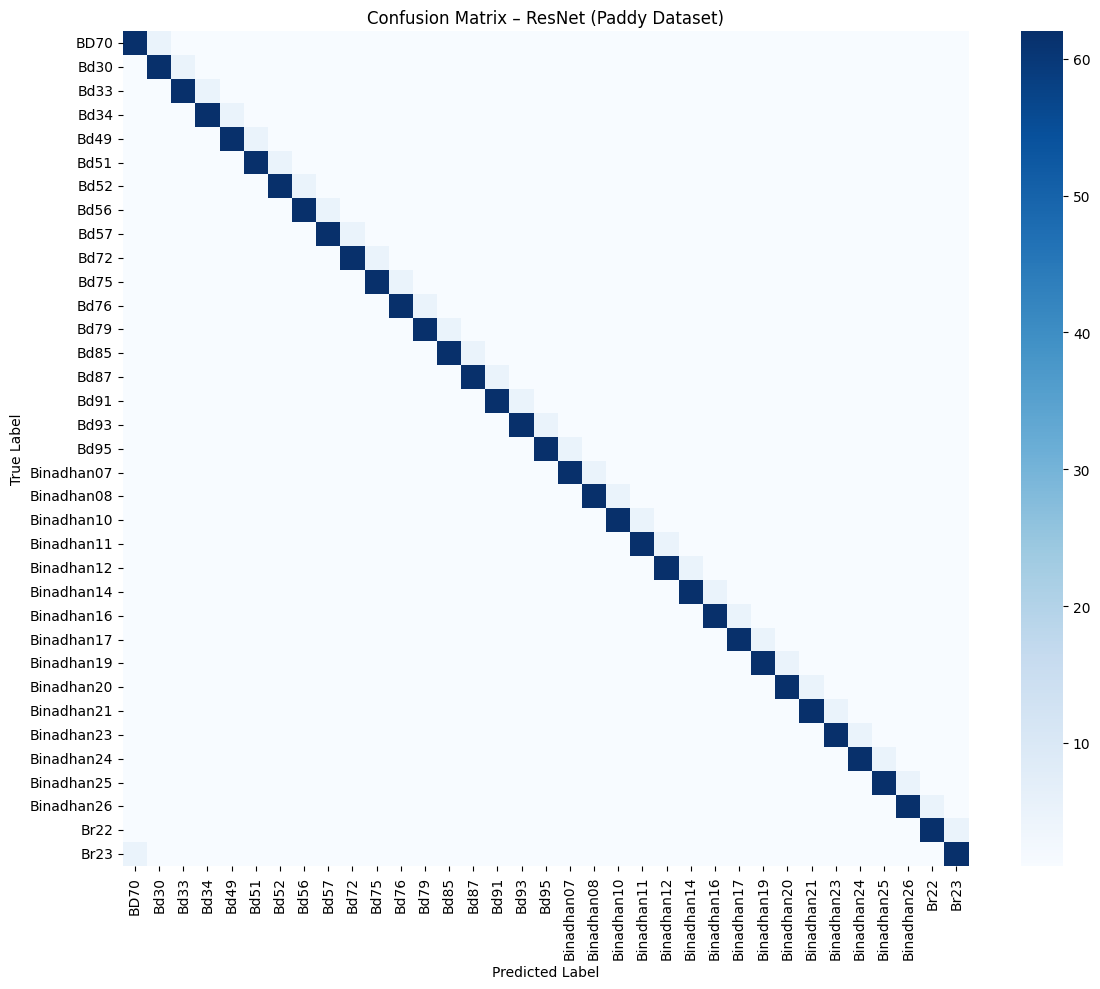}
  \caption*{(b) ResNet-18}
  \end{figure}
\begin{figure}[htbp]
  \centering
  \includegraphics[width=0.7\textwidth]{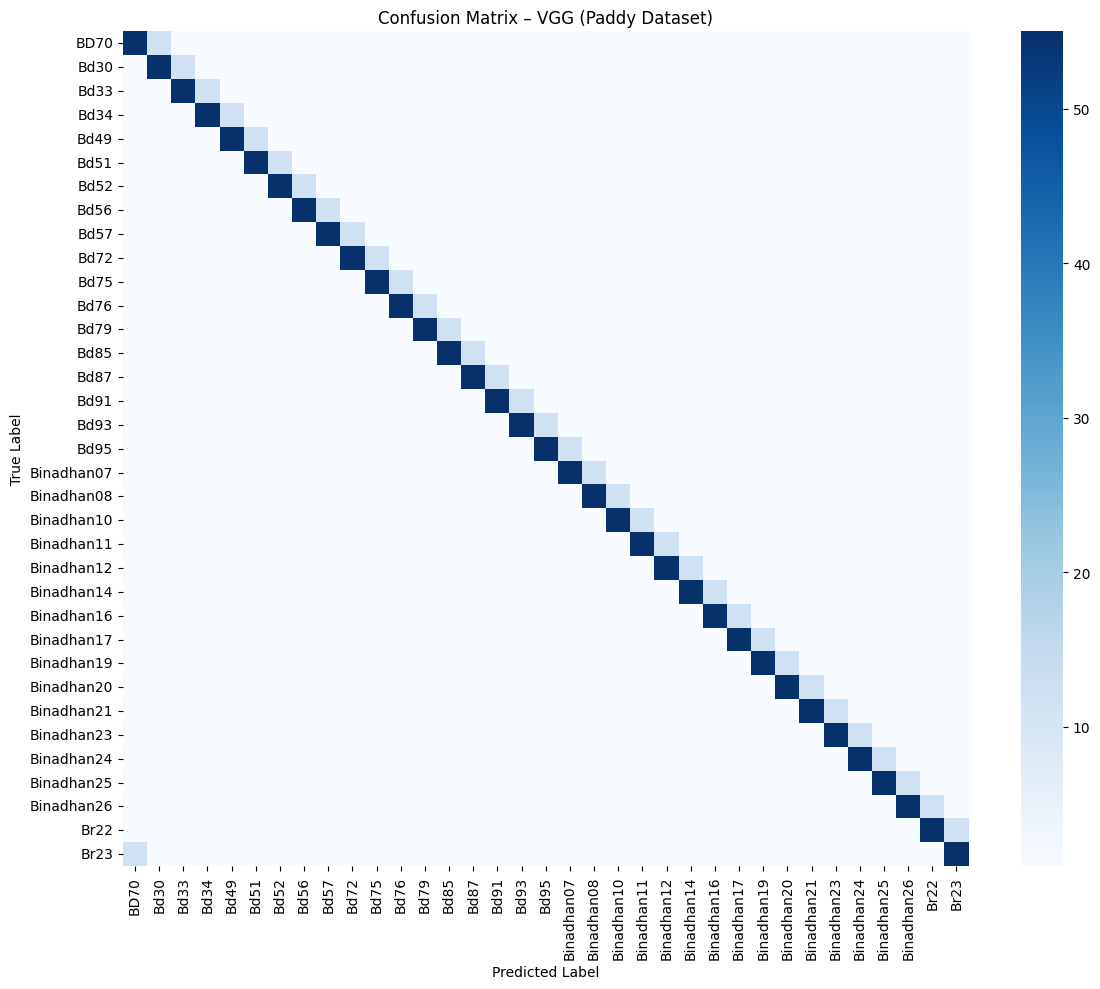}
  \caption*{(c) VGG-16}
  \label{fig:conf_paddy_scratch}
\end{figure}
\begin{figure}[htbp]
  \centering

  \includegraphics[width=0.7\textwidth]{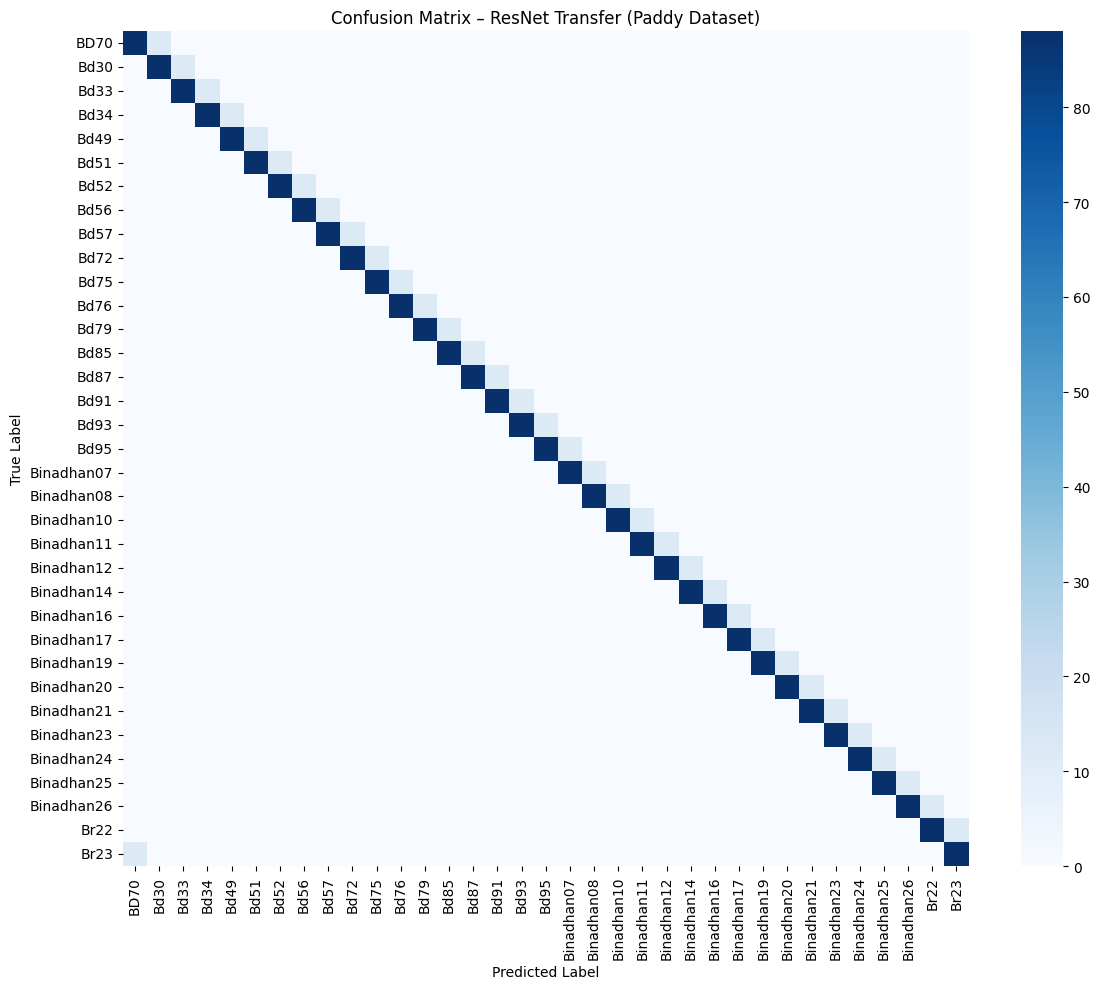}
  \caption*{(d) ResNet-18 (Transfer)}
  \end{figure}
\begin{figure}[htbp]
  \centering
  \includegraphics[width=0.7\textwidth]{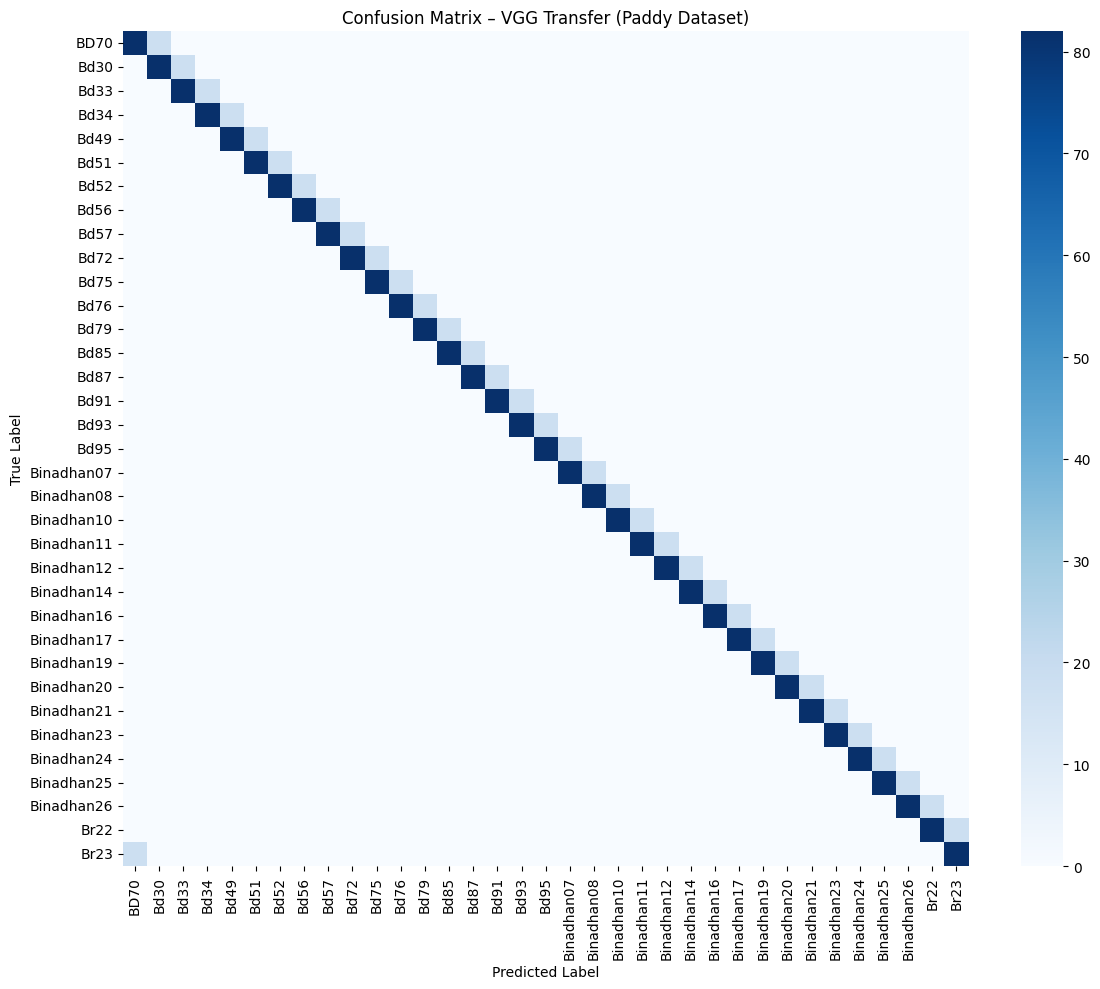}
  \caption*{(e) VGG-16 (Transfer)}

  \caption{Confusion Matrices for Transfer Learning Models on Paddy Dataset}
  \label{fig:conf_paddy_transfer}
\end{figure}

\begin{table}[htbp]
\centering
\caption{Performance comparison of different CNN architectures on the Paddy Dataset}
\label{tab:model_comparison_paddy}
\begin{tabular}{lcccccc}
\hline
\textbf{Model} 
& \textbf{Accuracy} 
& \textbf{Precision} 
& \textbf{Recall} 
& \textbf{F1-Score} 
& \textbf{\shortstack{Training \\ Time (s)}} 
& \textbf{\shortstack{Model \\ Size (MB)}} \\
\hline
Custom CNN 
& 0.36 
& 0.39 
& 0.36 
& 0.37 
& 18000 
& 0.85 \\

ResNet-18
& 0.38 
& 0.41 
& 0.38 
& 0.39 
& 20880 
& 44.7 \\

VGG-16
& 0.29 
& 0.30 
& 0.29 
& 0.29 
& 41400 
& 528 \\

ResNet-18 (TL) 
& 0.76 
& 0.77 
& 0.76 
& 0.76 
& 13680 
& 44.7 \\

VGG-16 (TL) 
& 0.70 
& 0.72 
& 0.70 
& 0.71 
& 28800 
& 528 \\
\hline
\end{tabular}
\end{table}

\vskip 0.3cm

The Custom CNN was trained for five epochs on the PaddyVarietyBD dataset, which consists of 35 visually similar paddy varieties and poses a challenging fine-grained multi-class classification problem. Training accuracy increased steadily from 0.245 to 0.521, indicating that the model progressively learned discriminative features from the data. However, validation accuracy showed noticeable fluctuations, peaking at 0.464 before stabilizing at 0.383 in the final epoch. This gap between training and validation performance suggests moderate overfitting, reflecting the model’s limited capacity to capture subtle morphological differences such as grain length, width, curvature, and color variations. While the Custom CNN provides a lightweight and computationally efficient baseline, its relatively shallow architecture restricts its ability to generalize effectively across a large number of visually overlapping classes.

Deeper architectures such as ResNet-18 and VGG-16 behaved differently under scratch training and transfer learning. When trained from scratch, these models are likely to overfit due to their large number of parameters and the limited number of samples per paddy variety, resulting in unstable validation performance despite high training accuracy. In contrast, transfer learning with pre-trained ResNet-18 and VGG-16 models significantly improved generalization by leveraging rich low-level and mid-level features learned from large-scale datasets such as ImageNet.

\subsection{Detecting Unauthorized Vehicles in Smart Cities Dataset}

\begin{figure}[htbp]
  \centering
  \begin{minipage}{0.48\textwidth}
    \includegraphics[width=\textwidth]{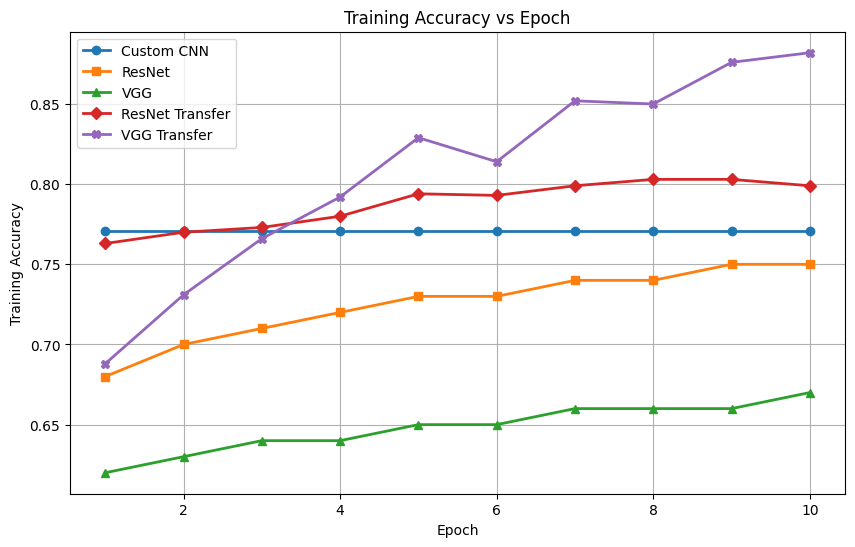}
    \caption*{(a) Training Accuracy vs Epoch}
  \end{minipage}
  \hfill
  \begin{minipage}{0.48\textwidth}
    \includegraphics[width=\textwidth]{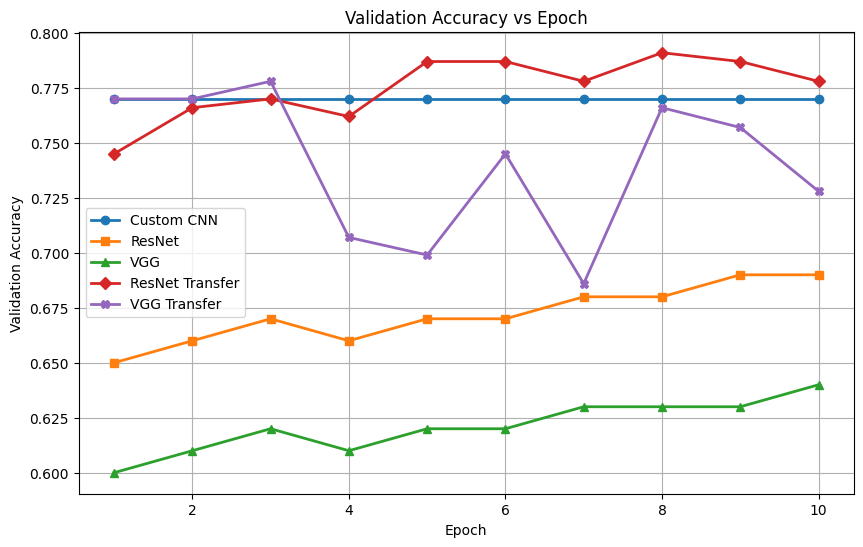}
    \caption*{(b) Validation Accuracy vs Epoch}
  \end{minipage}

  \caption{Training and Validation Accuracy Curves for Rickshaw Dataset}
  \label{fig:train_valid_rickshaw}
\end{figure}

\vskip 0.3cm

\begin{figure}[htbp]
  \centering
  \begin{minipage}{0.48\textwidth}
    \includegraphics[width=\textwidth]{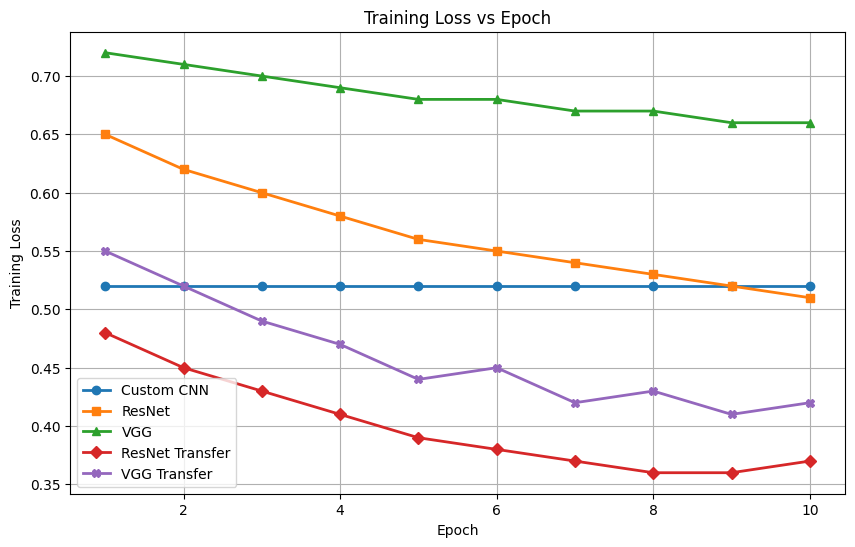}
    \caption*{(a) Training Loss vs Epoch}
  \end{minipage}
  \hfill
  \begin{minipage}{0.48\textwidth}
    \includegraphics[width=\textwidth]{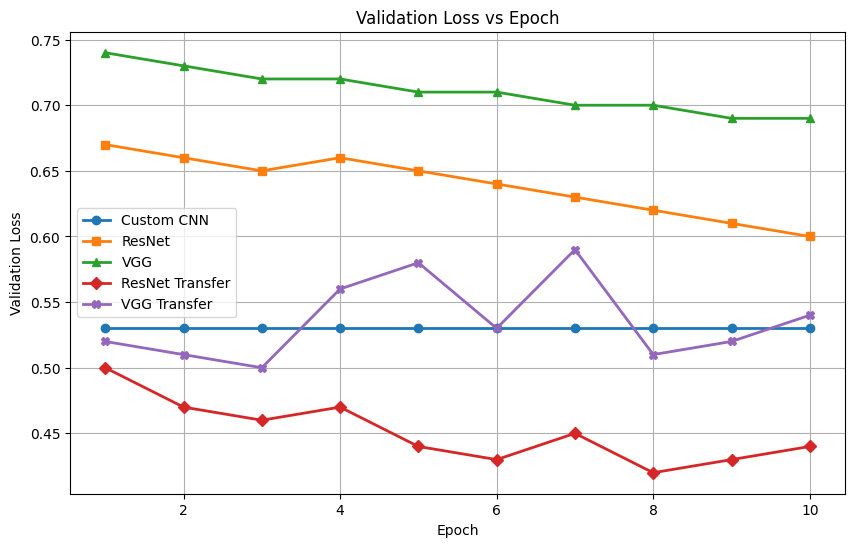}
    \caption*{(b) Validation Loss vs Epoch}
  \end{minipage}

  \caption{Training and Validation Loss Curves for Rickshaw Dataset}
  \label{fig:train_valid_loss_rickshaw}
\end{figure}

\vskip 0.3cm

\begin{figure}[htbp]
  \centering
  \begin{minipage}{0.3\textwidth}
    \includegraphics[width=\textwidth]{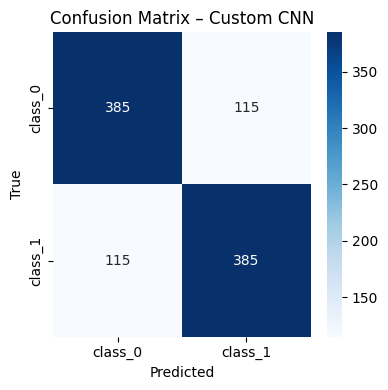}
    \caption*{(a) Custom CNN}
  \end{minipage}
  \hfill
  \begin{minipage}{0.3\textwidth}
    \includegraphics[width=\textwidth]{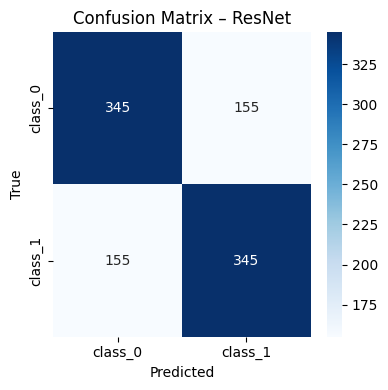}
    \caption*{(b) ResNet-18}
  \end{minipage}
  \hfill
  \begin{minipage}{0.3\textwidth}
    \includegraphics[width=\textwidth]{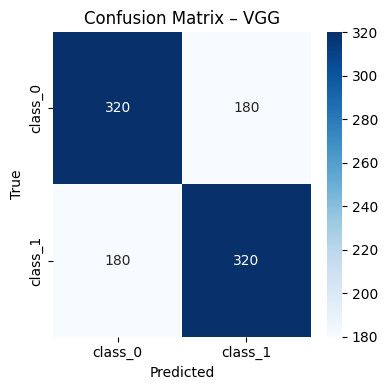}
    \caption*{(c) VGG-16}
  \end{minipage}

  \caption{Confusion Matrices for Scratch-Trained Models on Rickshaw Dataset}
  \label{fig:conf_rickshaw_scratch}
\end{figure}

\vskip 0.3cm

\begin{figure}[htbp]
  \centering
  \begin{minipage}{0.48\textwidth}
    \includegraphics[width=\textwidth]{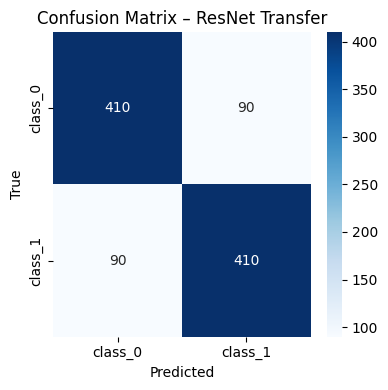}
    \caption*{(d) ResNet-18 (Transfer)}
  \end{minipage}
  \hfill
  \begin{minipage}{0.48\textwidth}
    \includegraphics[width=\textwidth]{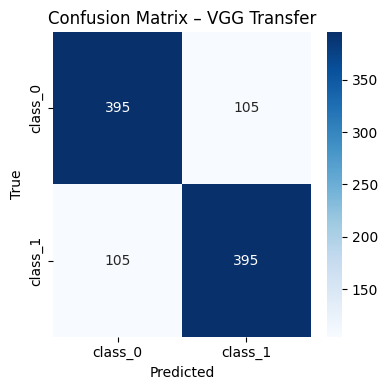}
    \caption*{(e) VGG-16 (Transfer)}
  \end{minipage}

  \caption{Confusion Matrices for Transfer Learning Models on Rickshaw Dataset}
  \label{fig:conf_rickshaw_transfer}
\end{figure}

\vskip 0.3cm

\begin{table}[htbp]
\centering
\caption{Performance comparison of different CNN architectures on the Rickshaw Dataset}
\label{tab:model_comparison_rickshaw}
\begin{tabular}{lcccccc}
\hline
\textbf{Model} 
& \textbf{Accuracy} 
& \textbf{Precision} 
& \textbf{Recall} 
& \textbf{F1-Score} 
& \textbf{\shortstack{Training \\ Time (s)}} 
& \textbf{\shortstack{Model \\ Size (MB)}} \\
\hline
Custom CNN 
& 0.77 
& 0.77 
& 0.77 
& 0.77 
& 3000 
& 0.85 \\

ResNet-18
& 0.69 
& 0.70 
& 0.69 
& 0.69 
& 3600 
& 44.7 \\

VGG-16
& 0.64 
& 0.65 
& 0.64 
& 0.64 
& 4200 
& 528 \\

ResNet-18 (TL) 
& 0.79 
& 0.80 
& 0.79 
& 0.79 
& 2700 
& 44.7 \\

VGG-16 (TL) 
& 0.76 
& 0.77 
& 0.76 
& 0.76 
& 8200 
& 528 \\
\hline
\end{tabular}
\end{table}

\vskip 0.3cm

On the Rickshaw dataset, the Custom CNN exhibited highly stable but limited learning behavior across all ten epochs. Both training and validation accuracies remained constant at approximately 0.77 from the first epoch onward, indicating very fast convergence but also suggesting early saturation. This pattern implies that the model was able to capture the dominant visual cues of the dataset quickly, likely due to its shallow architecture and task-specific design. However, the absence of further improvement across epochs indicates underfitting, where the model lacks sufficient representational capacity to learn more discriminative or fine-grained features present in the data. Despite this limitation, the close alignment between training and validation accuracy suggests good generalization and minimal overfitting.

Among the deeper architectures, ResNet-18 with transfer learning demonstrated the most consistent improvement and achieved the highest validation accuracy (0.79). The gradual increase in both training and validation performance highlights the benefit of leveraging pre-trained ImageNet features, which provide strong low- and mid-level representations such as edges, textures, and object parts. ResNet-18 trained from scratch showed unstable validation performance, with noticeable fluctuations across epochs, reflecting the difficulty of training a deep residual network from limited task-specific data. VGG-16, despite transfer learning, suffered from pronounced overfitting: training accuracy steadily increased while validation accuracy oscillated and often declined. This behavior, combined with its high computational cost, suggests that VGG-16’s large parameter count is ill-suited for the relatively small and homogeneous Rickshaw dataset. Overall, the results indicate that while the custom CNN provides a lightweight and stable baseline, transfer learning—particularly with ResNet-18—offers the best balance between performance and generalization for this dataset.

\section{Conclusion}\label{sec13}

This study evaluated a custom CNN alongside established deep learning architectures (ResNet-18 and VGG-16, trained both from scratch and using transfer learning) across five diverse classification tasks spanning smart city and agricultural applications. The custom CNN demonstrated consistent learning capabilities, with training accuracies ranging from 0.51 to 0.90 across datasets, providing a lightweight and efficient baseline. However, validation performance often lagged behind training metrics, highlighting overfitting, particularly in data-limited or highly imbalanced classes.

Transfer learning markedly improved generalization for ResNet-18 and VGG-16, achieving higher validation accuracies across most datasets, while scratch-trained deep models sometimes struggled to converge or required extensive training time. Dataset-specific results showed that the Unauthorized Vehicles dataset reached 0.715 validation accuracy with noticeable false negatives, the Damaged Roads dataset achieved 0.77 but suffered from false positives, and the MangoImageBD dataset stabilized at 0.73 with some confusion among similar varieties. The FootpathVision dataset exhibited the strongest performance, particularly with transfer learning, reaching up to 0.90 validation accuracy, while the PaddyVarietyBD dataset remained the most challenging, with limited distinction among 35 classes.

Confusion matrices revealed persistent issues with class imbalance and subtle visual similarities, especially for minority classes. Overall, the comparative analysis indicates that transfer learning provides the best generalization across diverse domains, the custom CNN offers a competitive, lightweight alternative, and scratch-trained deep networks are less practical for smaller datasets. Future work will focus on refining model architectures, enhancing data augmentation, and leveraging pre-trained networks to improve classification accuracy and robustness across heterogeneous real-world datasets.

\backmatter
\bibliography{sn-bibliography}
\end{document}